\ifdefined\pdfminorversion\pdfminorversion=7\fi
\RequirePackage{fix-cm}
\documentclass[10pt]{lynnreal}
\usepackage{lmodern}
\usepackage{amsmath,amssymb,booktabs,tikz,array}
\usepackage{xurl}
\usepackage{needspace}
\usepackage{adjustbox}
\usetikzlibrary{arrows.meta}
\microtypesetup{expansion=true}
\hypersetup{
  pdftitle={LynnReal-Hand: Unlocking World Interaction with Hand Control},
  pdfauthor={Shaohao Rui, Chaoyang Zhang, Xiaofeng Mao, Weijie Ma},
  pdfsubject={Technical report on hand-controlled video world modeling}
}
\title{\adjustbox{max width=\linewidth}{\LARGE LynnReal-Hand: Unlocking World Interaction with Hand Control}}

\author[*]{LynnReal AI}
\abstract{Interactive video world models enable exploration through text and camera controls, yet direct object interaction requires finer control. Hands provide a natural interface, but translating continuous hand movements into coherent visual responses remains challenging because of complex hand-object dynamics and demanding real-time computation. To address these challenges, we introduce LynnReal-Hand, a hand-controlled video world model that unifies precise interaction modeling with efficient streaming generation. We curate 100K interaction clips spanning diverse real and synthetic environments and develop hand-aware supervision to improve fine-grained hand motions and object interactions. Building on this foundation, we learn a causal flow map for few-step generation and introduce region-aware score distillation that combines complementary video and image priors to refine interaction dynamics and visual appearance. To further reduce inference latency, we propose Ultra-Fast-VAE, a compact video decoder that preserves high reconstruction fidelity with substantially lower computational cost. Extensive evaluations on HandBench demonstrate superior hand-following accuracy and visual quality compared with existing approaches. Our full-sequence model achieves an FVD of 181.67 and a MANO Dice score of 93.02, while the three-step causal model maintains a MANO Dice score of 92.49. For real-time interaction, 540p streaming on a single H200 reaches 18.03 FPS with 0.741\,s latency and an optimized peak GPU memory footprint of 34.89\,GiB. We will release the datasets, model weights, and the full training and inference code to support research on hand-controlled video world models.}
\date{Technical report, November 17, 2026}
\lynndata[Code]{\url{https://github.com/LynnReal-AI/LynnReal}}
\lynndata[Models]{\url{https://huggingface.co/shaohao011/LynnReal-Hand-Base-1.0}\par
\hspace*{3.6em}\url{https://huggingface.co/shaohao011/LynnReal-Hand-Fast-preview}\par
\hspace*{3.6em}\url{https://huggingface.co/shaohao011/LynnReal-ultra-fast-vae-1.0}}

\begin{document}
\vspace*{-12pt}
\maketitle
\begingroup
\renewcommand{\thefootnote}{\fnsymbol{footnote}}
\footnotetext[1]{Contributors and affiliations are listed in Appendix~\ref{sec:contributors}.}
\endgroup
\IfFileExists{fig/teaser-v2.jpg}{%
\vspace{-5pt}
\begin{center}
\captionsetup{font=small,skip=4pt,hypcap=false}
\includegraphics[width=\textwidth]{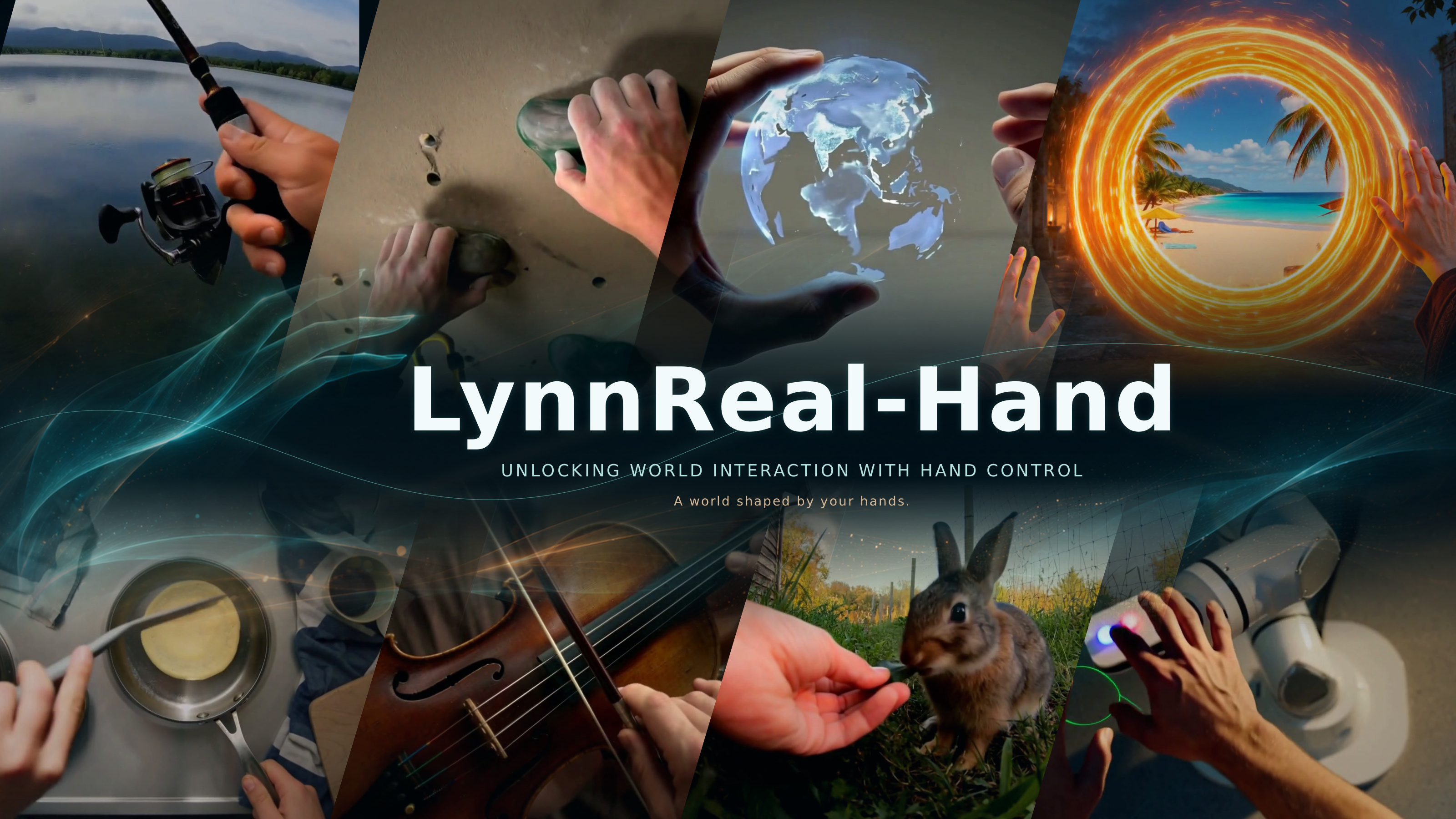}
\captionof{figure}{Hand-conditioned world interaction across everyday and imaginative scenes.}
\label{fig:teaser}
\end{center}
}{}
\clearpage
\hyphenpenalty=50
\exhyphenpenalty=50
\tolerance=2000
\emergencystretch=3em

\setlength{\textfloatsep}{10pt plus 2pt minus 2pt}
\setlength{\floatsep}{8pt plus 2pt minus 2pt}
\setlength{\intextsep}{8pt plus 2pt minus 2pt}
\setlength{\dbltextfloatsep}{12pt plus 2pt minus 2pt}
\setlength{\dblfloatsep}{8pt plus 2pt minus 2pt}
\renewcommand{\topfraction}{.92}
\renewcommand{\dbltopfraction}{.92}
\renewcommand{\textfraction}{.06}
\renewcommand{\floatpagefraction}{.88}
\renewcommand{\dblfloatpagefraction}{.88}
\captionsetup{font=small,skip=4pt}
\makeatletter
\setlength{\@fptop}{0pt}
\setlength{\@fpsep}{12pt}
\setlength{\@fpbot}{0pt plus 1fil}
\makeatother

\section{Introduction}
\label{sec:introduction}

Interactive video world models are transforming video generation from passive content synthesis into environments that respond to user actions. Early systems such as Genie~\cite{genie} demonstrated action-controllable generation, while recent advances in Matrix-Game 3.0~\cite{matrix3}, PAN~\cite{pan}, and open world models~\cite{worldplay,relic,lingbot,solarwm} have improved interactivity, temporal consistency, and accessibility. Yet most interaction paradigms remain centered on navigation, camera motion, or high-level instructions. These controls determine where users look or move, but offer limited ability to manipulate the objects they encounter. Enabling such manipulation requires a world model to respond not only to changes in viewpoint, but also to the fine-grained actions through which users interact with their surroundings.

Hands offer a direct and expressive interface for this purpose: reaching, grasping, releasing, and coordinated bimanual movements communicate intentions that are difficult to specify through text or navigation commands. Recent efforts, including HandWorld~\cite{handworld}, Hand2World~\cite{hand2world}, Generated Reality~\cite{generatedreality}, Dexterous World Models~\cite{dwm}, and JointControlVideo~\cite{jointcontrolvideo}, have begun to connect articulated hand motion with visual generation. However, reliable hand-controlled interaction presents challenges that are distinct from conventional video synthesis. First, hands and their contact regions occupy only a small portion of each frame, so global generative objectives can underemphasize precisely the details that determine control fidelity. Second, accelerating a high-quality video model into a few-step causal generator can amplify local motion errors: the student must preserve articulated interactions while conditioning on its own imperfect history. Finally, even when denoising is accelerated, decoding high-resolution video latents can remain a major obstacle to responsive interaction. These challenges call for a training and inference pipeline that preserves interaction fidelity throughout the transition from full-sequence learning to streaming generation.

We introduce \textbf{LynnReal-Hand}, an open-source hand-controlled video world model developed to address these challenges end to end. At its core is \textbf{Interaction-Preserving Training (IPT)}, a supervision strategy that explicitly prioritizes fine-grained interaction fidelity at two critical stages: full-sequence pretraining and causal distribution matching. We first build a dataset of \textbf{100K interaction clips} with aligned hand trajectories, camera poses, and textual descriptions across diverse real and synthetic scenes. During full-sequence learning, we go beyond global flow matching with \emph{hand-aware multilevel supervision}: a hand-region latent objective penalizes local reconstruction errors directly, while decoded LPIPS and DINOv2 constraints reinforce perceptual structure and hand appearance. This targeted supervision counteracts the weak contribution of small interaction regions to the global training signal and establishes a stronger teacher for subsequent acceleration.

Building on this pretrained model, we learn a \textbf{causal flow map}~\cite{anyflow} that generates each video chunk in only three denoising transitions. Teacher-forced training with clean history provides an efficient initialization, but leaves a mismatch with inference, where each new chunk depends on previously generated content. We therefore refine the causal student on self-generated rollouts using distribution matching~\cite{dmd,selfforcing}. However, global distribution matching alone does not distinguish the visual evidence needed for interaction dynamics from the evidence needed for scene appearance. To allocate these complementary signals, we formulate \emph{region-aware dual-score distillation}: a hand-conditioned video teacher and critic guide the entire generated sequence, with higher-noise evaluations emphasizing interaction-related corrections, while a complementary image prior supplies appearance guidance only outside hand regions. This spatial separation is essential because the image prior lacks hand conditions and should not override the specified motion. In this way, IPT addresses complementary failure modes of the two training stages: insufficient hand-local learning during pretraining and degraded interaction fidelity during causal acceleration. For open-ended streaming, a bounded attention window and local temporal-position reset keep the student's history and positional range finite.

Efficient generation also requires efficient reconstruction. We therefore introduce Ultra-Fast-VAE, a compact Transformer decoder that preserves the generator's native latent interface while compressing attention and feed-forward branches and recovering reconstruction quality through representation and temporal distillation. We evaluate the resulting system with HandBench, covering ten hand-interaction sources, including two held-out sources for out-of-distribution testing. On the main benchmark, LynnReal-Hand-Base achieves an FVD of 181.67 and a MANO Dice score of 93.02; the evaluated LynnReal-Hand-Fast model retains 92.49 MANO Dice with three denoising steps. At 540p on one H200, Fast achieves 0.741\,s latency and 18.03 FPS.

Our main contributions are summarized as follows:
\begin{itemize}
    \item Interaction-scale data and evaluation. We curate 100K hand-interaction clips with temporally aligned hand, camera, and semantic conditions, and establish HandBench to assess control fidelity and visual quality across diverse scenes and held-out settings.

    \item Interaction-Preserving Training. We develop a unified supervision perspective spanning hand-aware multilevel learning for full-sequence generation and region-aware score allocation for causal distillation, explicitly addressing how hand-control signals are preserved at both stages.

    \item Few-step, open-ended causal generation. We combine causal flow-map learning, refinement on self-generated histories, and bounded-window temporal-position management to support three-step streaming generation without an ever-growing attention cache.

    \item Fast, high-fidelity video decoding. We propose Ultra-Fast-VAE, which compresses a pretrained Transformer decoder while retaining its native latent interface and restoring spatial and temporal reconstruction fidelity through targeted distillation.
\end{itemize}

\section{Related Work}
\label{sec:related}

\subsection{Interactive Video World Models}
\label{sec:related_world}
Video world models simulate environment dynamics conditioned on user inputs, enabling interactive exploration and planning. Recent systems have scaled action-conditioned synthesis, geometric consistency, and long-horizon memory across open-domain and gaming environments~\cite{genie,matrix3,pan,worldplay,relic,lingbot,solarwm,biwm,mao2026yume1}. Their controls mainly specify camera trajectories, discrete actions, or text prompts, with limited spatial precision for object manipulation. We use articulated 3D hand motion as the control signal and design supervision and causal inference around hand--object interaction.

\subsection{Hand-Controlled Video Generation}
\label{sec:related_hand}
To enable dexterous manipulation, recent works condition video generators on projected hand meshes, skeletal joints, or structured action representations~\cite{handworld,hand2world,generatedreality,dwm,dexac,p3wam,jointcontrolvideo,dreamdojo,egowm}. While some approaches jointly model hand trajectories and scene dynamics~\cite{handworld,hand2world,dreamdojo} or combine hand inputs with camera and static-scene priors~\cite{generatedreality,dwm,jointcontrolvideo}, they largely rely on spatially uniform training objectives and lack open, end-to-end streaming pipelines. Because hands occupy a small fraction of each frame, uniform objectives under-supervise finger articulation, while naive image-level guidance during distillation conflicts with prescribed hand poses. Our IPT directly addresses both bottlenecks by enforcing multi-level local hand constraints during pretraining and decoupling hand-aware video guidance from background-only image guidance during causal distillation.

\subsection{Efficient Video Generation and Decoding}
\label{sec:related_efficient}
Accelerating video generation requires reducing both sampling steps and per-frame decoding overhead. Few-step flow-map learning~\cite{anyflow} and distribution-matching distillation~\cite{dmd,duomatching}, alongside autoregressive self-rollout training~\cite{selfforcing,packforcing}, enable low-latency causal generation without severe error accumulation. Yet in hand-controlled settings, aggressive step reduction degrades small, fast-moving fingers, while high-capacity VAE decoders remain a major runtime bottleneck—and naive decoder compression blurs fine manipulation details. Rather than accelerating denoising and decoding in isolation, we pair IPT-guided causal distillation with Ultra-Fast-VAE to preserve fine finger fidelity end-to-end.

\begin{figure}[!htbp]
\centering
\includegraphics[width=.90\textwidth]{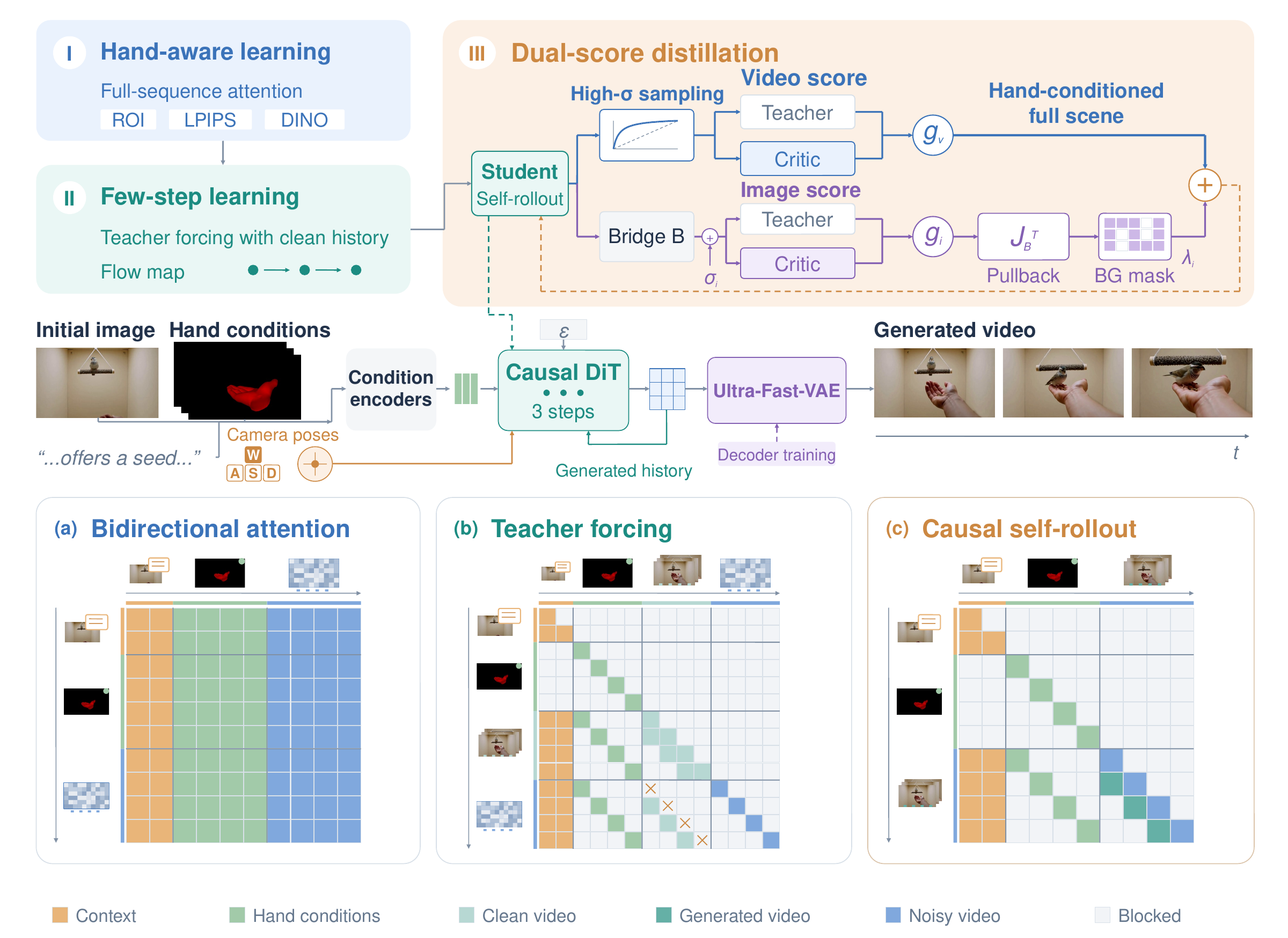}
\caption{Overview of LynnReal-Hand. Stage I learns a hand-aware bidirectional model; Stage II adapts it to few-step causal generation with clean history; Stage III distills self-rollouts using high-noise video and background-only image guidance. Ultra-Fast-VAE supports streaming decoding. In attention masks (a)--(c), rows/columns are queries/keys; colored blocks allow attention and gray/$\times$ blocks mask it.}
\label{fig:pipeline}
\end{figure}

\section{Methods}
\label{sec:method}

Interaction-Preserving Training (IPT) learns the generator through hand-aware pretraining, causal flow-map adaptation, and region-aware dual-score distillation on student rollouts. Ultra-Fast-VAE then decodes each generated latent chunk at low latency.

\subsection{Hand-Controlled Video Generation}
\label{sec:formulation}

Our goal is to synthesize a temporally coherent video $X$ conditioned on an initial frame $I_0$, a text prompt $c$, hand motion signals $H$, and camera poses $P$. Grouping these multimodal conditions as $\mathcal{C} = (I_0, c, H, P)$, we formulate the task in the latent space of a pretrained encoder $z = \mathcal{E}(X)$ using Conditional Flow Matching (CFM)~\cite{flowmatching}.

Specifically, we define a linear interpolation path between the clean video latent $z$ and Gaussian noise $\epsilon \sim \mathcal{N}(0, I)$:
\begin{equation}
    z_\sigma = (1 - \sigma)z + \sigma\epsilon, \qquad v^\star = \epsilon - z,
    \label{eq:flowpath}
\end{equation}
where $\sigma \in [0, 1]$ is the noise level ($0$ is clean and $1$ is pure noise), and $v^\star$ is the target velocity field. A velocity predictor $v_\theta(z_\sigma, \sigma; \mathcal{C})$ is trained via the standard flow matching objective:
\begin{equation}
    \mathcal{L}_{\mathrm{FM}} = \mathbb{E}_{z, \epsilon, \sigma} \left[ \left\| v_\theta(z_\sigma, \sigma; \mathcal{C}) - v^\star \right\|_2^2 \right].
    \label{eq:stage1loss}
\end{equation}
While $\mathcal{L}_{\mathrm{FM}}$ effectively captures global scene and camera dynamics, its spatially uniform weighting provides insufficient supervision for small, articulated hand regions, motivating our hand-aware training strategy below.

\subsection{Stage I: Hand-Aware Pretraining}
\label{sec:stageone}

To strengthen supervision in small hand regions, Stage~I enforces local constraints across latent, perceptual, and semantic spaces. Given the predicted clean latent $\widehat z=z_\sigma-\sigma v_\theta(z_\sigma,\sigma;\mathcal C)$ and a latent-aligned binary hand mask $M_H$, we add three hand-localized losses:
\begin{equation}
    \mathcal{L}_{\mathrm{I}} = \mathcal{L}_{\mathrm{FM}} + \lambda_R \mathcal{L}_{\mathrm{ROI}} + \lambda_P \mathcal{L}_{\mathrm{LPIPS}}^H + \lambda_D \mathcal{L}_{\mathrm{DINO}}^H,
    \label{eq:stage1handloss}
\end{equation}
where $\mathcal{L}_{\mathrm{ROI}} = \| M_H \odot (\widehat{z} - z) \|_2^2$ directly penalizes latent reconstruction errors within hand regions. To further preserve fine finger articulation and structural semantics in pixel space, $\mathcal{L}_{\mathrm{LPIPS}}^H$ and $\mathcal{L}_{\mathrm{DINO}}^H$ compute perceptual~\cite{lpips} and patch-level semantic~\cite{dinov2} distances between cropped hand regions of the decoded prediction $\mathcal{D}(\widehat{z})$ and ground truth, using frozen feature backbones.

\textbf{Architecture and Conditioning.}
As shown in Fig.~\ref{fig:pipeline}, we build upon a bidirectional Diffusion Transformer (DiT)~\cite{dit} operating in the latent space of the frozen H3 VAE~\cite{minimaxh3}. Text instructions $c$ and the initial frame $I_0$ are encoded via Qwen3-VL~\cite{qwen3vl}, while hand motions $H$ are tokenized by the H3 VAE. These conditioning tokens are concatenated with the noisy video latents along the sequence dimension, and camera poses $P$ are injected via Projective Positional Encoding (PRoPE)~\cite{prope}. We optimize all DiT parameters while keeping all conditioning encoders frozen. The resulting bidirectional model serves as both the initialization for causal adaptation (Stage~II) and the frozen hand-aware video teacher for distillation (Stage~III). Loss weights and training details are in Appendix~\ref{sec:supp_stage1_training}.

\subsection{Stage II: Causal Flow-Map Adaptation}
\label{sec:stagetwo}

While Stage~I achieves high hand fidelity, its bidirectional attention and multi-step sampling incur prohibitive latency for interactive world modeling. To enable low-latency streaming, we adapt the bidirectional model via LoRA~\cite{lora} into a few-step causal generator using AnyFlow~\cite{anyflow} and a two-chunk causal window ($\mathrm{W2}$).

\textbf{Few-step flow-map learning.}
For chunk $k$, let $\mathcal{C}_k$ bundle the text/image context, aligned hand and camera controls, and the preceding clean history chunk. AnyFlow learns a direct jump predictor $U_\theta$ that maps a noisy state $z_{k,\sigma}$ at noise level $\sigma$ to a cleaner state $z_{k,r}$ ($r < \sigma$):
\begin{equation}
    z_{k,r} = z_{k,\sigma} - (\sigma - r) U_\theta(z_{k,\sigma}, \sigma, r; \mathcal{C}_k).
    \label{eq:anyflow_step}
\end{equation}
Here a sampling step denotes one flow-map update between two noise levels; three-step sampling traverses three such intervals.
$U_\theta$ estimates average velocity over $[r, \sigma]$ and is trained via
\begin{equation}
    \mathcal{L}_{\mathrm{II}} = \mathbb{E} \left[ \left\| U_\theta + (\sigma - r)\widehat{D}_\sigma U_\theta - v^\star \right\|_M^2 \right],
    \label{eq:anyflow_loss}
\end{equation}
where $\widehat{D}_\sigma U_\theta$ denotes the stop-gradient finite-difference approximation of the total derivative along the trajectory (reducing to standard flow matching when $r=\sigma$). Here $M$ is the binary validity mask over target-video latent entries, excluding conditioning, clean-history, and padding tokens; $\|a\|_M^2=\sum_j M_j a_j^2/\max(\sum_j M_j,1)$ is their mean-squared error. Inference uses three transitions per chunk.

\textbf{Two-chunk causal masking.}
As illustrated in Fig.~\ref{fig:pipeline}(b), we decouple clean-history and noisy-target streams under a $\mathrm{W2}$ sliding window (comprising the current chunk and one preceding chunk at full spatial resolution). During training, clean chunks attend causally to themselves and the immediate predecessor, while noisy target chunks attend to the shared conditions, the preceding clean chunk, and themselves—masking out their own clean ground truth and future chunks to prevent information leakage. Masking and schedule-specific AdaLN caching are detailed in Appendices~\ref{sec:supp_attention_rope} and~\ref{sec:supp_inference_variants}, respectively.

\begin{figure}[!htbp]
\centering
\includegraphics[width=.98\textwidth]{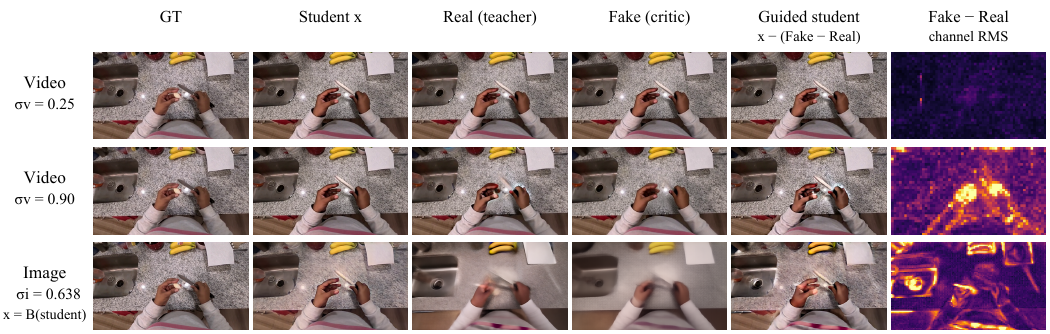}
\caption{\textbf{Real/fake model predictions before DMD normalization and hand masking.} Columns: GT, student $x$, real-model prediction $D^r$, fake-model prediction $D^f$, unit-step output $x+(D^r-D^f)$, and channel-RMS prediction difference $\|D^r-D^f\|_{\mathrm{RMS}}$. High-noise video guidance ($\sigma_v=0.90$, middle) concentrates corrections on hands and objects compared with low noise ($0.25$, top). The hand-blind image branch ($\sigma_i=0.638$, bottom) sharpens backgrounds but degrades hand poses, motivating Eq.~\eqref{eq:region_aware_direction}.}
\label{fig:score_responses}
\end{figure}

\subsection{Stage III: Region-Aware Distillation}
\label{sec:duo}

Since Stage~II relies solely on ground-truth histories, autoregressive rollouts inevitably suffer from exposure bias and compounding errors. Video-only self-rollout distillation~\cite{selfforcing,dmd} reduces temporal drift but can lose spatial detail. Inspired by DuoMatching~\cite{duomatching}, we augment the video teacher with an auxiliary image teacher to restore high-frequency appearance details. A hand-blind image teacher can produce gradients that conflict with prescribed finger poses. Adapting DuoMatching to preserve fine hand fidelity, \emph{region-aware dual-score distillation}---the second component of IPT (Fig.~\ref{fig:pipeline})---decouples motion and background guidance by noise level and spatial region.

During training, the three-step causal student generates rollout chunks $Z_\theta$ conditioned on its own detached preceding predictions under the $\mathrm{W2}$ mask, while a local RoPE reset bounds temporal positions across sliding windows without altering relative camera poses (Appendix~\ref{sec:supp_attention_rope}).

\textbf{High-noise-shifted video guidance.}
As shown in Fig.~\ref{fig:score_responses} (top), evaluating the video branch at low noise ($\sigma_v = 0.25$) yields nearly identical teacher and critic predictions, resulting in negligible score-difference gradients. To enforce accurate hand kinematics, we shift video evaluation toward higher noise levels by corrupting the student rollout via $\widetilde{Z}_v = (1 - \tau_v)Z_\theta + \tau_v \epsilon_v$ under a warped noise schedule:
\begin{equation}
    \tau_v = \frac{\gamma \sigma_v}{1 + (\gamma - 1)\sigma_v}, \qquad \gamma > 1,
    \label{eq:noise_warp}
\end{equation}
while keeping the student's three-step schedule unchanged. At higher noise ($\sigma_v = 0.90$, Fig.~\ref{fig:score_responses}, middle), the video branch suppresses minor texture discrepancies and concentrates corrective gradients on articulated hands and interacted objects, yielding clean predictions $D_v^r$ (teacher) and $D_v^f$ (critic).

\textbf{Reverse-KL distribution matching.}
We map rollout frames into the frozen Qwen-Image~\cite{qwenimage} teacher's latent space with the differentiable bridge $Y_\theta=B(Z_\theta,o)$ at a sampled frame offset $o$. Write $X_v=Z_\theta$ and $X_i=Y_\theta$ for the video and image branches. Each branch evaluates a perturbed sample $\widetilde X_b=(1-\tau_b)X_b+\tau_b\epsilon_b$, with its own noise level $\tau_b$.

For each branch, DMD~\cite{dmd} minimizes the reverse KL divergence from the noised student distribution $q^\theta_{b,\tau_b}$ to the target distribution represented by the frozen real model $p^r_{b,\tau_b}$:
\begin{subequations}
\label{eq:dual_branch_dmd}
\begin{equation}
 \mathcal L_{\mathrm{DMD}}^b
 =\mathbb E_{\tau_b}\!\left[
 w_b(\tau_b)\,
 D_{\mathrm{KL}}\!\left(
 q^\theta_{b,\tau_b}(\cdot\mid\mathcal C_b)
 \,\|\,p^r_{b,\tau_b}(\cdot\mid\mathcal C_b)
 \right)\right],\qquad b\in\{v,i\}.
 \label{eq:branch_kl}
\end{equation}
The video condition $\mathcal C_v$ includes the prescribed hands and cameras together with the shared context; the image condition $\mathcal C_i$ contains no hand or camera controls. In each branch, the real model is the frozen teacher, while the fake model $p^f_{b,\tau_b}$ is a critic trained on detached student samples to approximate $q^\theta_{b,\tau_b}$. Their video and image scores are
\begin{equation}
\begin{aligned}
 s_v^r&=\nabla_{\widetilde Z_v}\log p^r_{v,\tau_v}(\widetilde Z_v\mid\mathcal C_v),
 &s_v^f&=\nabla_{\widetilde Z_v}\log p^f_{v,\tau_v}(\widetilde Z_v\mid\mathcal C_v),\\
 s_i^r&=\nabla_{\widetilde Y_i}\log p^r_{i,\tau_i}(\widetilde Y_i\mid\mathcal C_i),
 &s_i^f&=\nabla_{\widetilde Y_i}\log p^f_{i,\tau_i}(\widetilde Y_i\mid\mathcal C_i).
\end{aligned}
 \label{eq:branch_scores}
\end{equation}
The video scores are evaluated on full-sequence latents, and the image scores on bridge-projected frame latents. DMD estimates the reverse-KL gradient from the real and fake score difference:
\begin{equation}
 \nabla_\theta\mathcal L_{\mathrm{DMD}}^b
 \simeq
 -\mathbb E\!\left[
 w_b(\tau_b)\,
 J_{\widetilde X_b,\theta}^{\top}
 \operatorname{sg}(s_b^r-s_b^f)
 \right],
 \label{eq:branch_kl_gradient}
\end{equation}
where $J_{\widetilde X_b,\theta}$ is the Jacobian of the perturbed sample with respect to the student parameters. The real and fake model scores are detached during this update; the fake model is optimized separately on student-generated samples.

For our perturbation, clean predictions and score estimates satisfy
$s_b^{r/f}=((1-\tau_b)D_b^{r/f}-\widetilde X_b)/\tau_b^2$ for $0<\tau_b<1$.
We absorb positive timestep factors into branch weighting and normalize the reverse-KL gradient per sample:
\begin{equation}
 g_b
 =\frac{\tau_b^2(s_b^f-s_b^r)}{(1-\tau_b)N_b}
 =\frac{D_b^f-D_b^r}{N_b},\qquad
 N_b=\max\!\left(\operatorname{mean}|X_b-D_b^r|,\delta\right).
 \label{eq:distillation_directions}
\end{equation}
\end{subequations}
Here $\delta>0$ prevents division by zero. The image contribution is pulled back through the frozen bridge before applying the hand mask; the video contribution remains global.

\textbf{Hand-masked image guidance.}
The hand-blind image real model sharpens backgrounds but can alter hand poses (Fig.~\ref{fig:score_responses}, bottom), producing large real/fake prediction differences in both regions. We therefore mask the hand region $M_H$ after pulling the image DMD gradient back through the bridge $B$, while retaining the video DMD gradient over the full scene:
\begin{equation}
    g_{\mathrm{RA}} = g_v + \lambda_i (1 - M_H) \odot J_B^\top g_i,
    \label{eq:region_aware_direction}
\end{equation}
where $J_B = \partial B / \partial Z_\theta$. Hand regions receive guidance from the control-aware video branch, while the image branch supplies background appearance gradients.

\textbf{Optimization.}
We update the student by minimizing the stop-gradient surrogate objective:
\begin{equation}
    \mathcal{L}_{\mathrm{III}} = \frac{1}{2} \mathbb{E} \left[ \left\| Z_\theta - \operatorname{sg}(Z_\theta - g_{\mathrm{RA}}) \right\|_2^2 \right],
    \label{eq:stage3loss}
\end{equation}
alternating with critic denoising on detached rollouts; $\lambda_i$ ramps up during warmup (Appendix~\ref{sec:supp_stage3_training}).

\section{Ultra-Fast-VAE: Architecture and Training}
\label{sec:vae}

Few-step distillation accelerates latent synthesis, but video decoding remains a bottleneck. Transformer decoders are costly even with streaming optimizations~\cite{streamrender,solarxvae}, whereas Tiny-H3, the H3-compatible lightweight CNN autoencoder from TAEHV~\cite{taehv}, blurs fine finger and object detail. Ultra-Fast-VAE addresses this trade-off with cache-free, width-compressed Transformer decoding and boundary-aware distillation in the native latent space.

\textbf{Structured width compression with ridge initialization.}
Rather than uniformly shrinking network depth or residual dimensions, which degrades representational capacity, we preserve the full backbone depth (26 blocks) and 2048-dimensional residual stream of the pretrained H3-compatible LightVAE~\cite{lynnrealomni}, while aggressively pruning intra-block computation (reducing attention heads from 32 to 8 and SwiGLU~\cite{swiglu} hidden width from 8192 to 1536). To retain the most informative subspaces, we score each intermediate channel $j$ via its activation-weighted output norm:
\begin{equation}
    s_j = \mathbb{E}[a_j^2] \left\| \gamma \odot W_{:,j} \right\|_2^2,
    \label{eq:vae_importance}
\end{equation}
where $a_j$ is the channel activation, $W_{:,j}$ is the downstream projection column, and $\gamma$ denotes the residual scaling factor. FFN channels are pruned individually, while attention heads are ranked by aggregating channel scores within each head.

Since naive slicing introduces truncation errors, we recalibrate each compressed output projection $(B, b)$ via ridge regression to match the full-width branch output $Y$:
\begin{equation}
 \min_{B,b}\;\|A_SB+\mathbf1b^\top-Y\|_F^2
 +\lambda_W\|B-B_0\|_F^2+\lambda_b\|b-b_0\|_2^2.
 \label{eq:vae_ridge}
\end{equation}
where $A_S$ denotes the selected activations and $(B_0, b_0)$ are the sliced weights. This closed-form initialization preserves the primary linear mapping prior to fine-tuning, enabling cache-free decoding without lookahead.

\textbf{Operator distillation and hand-aware temporal continuity.}
Reconstruction training matches the full-width decoder's operators and image outputs to recover detail after compression. Instead of matching intermediate features offline, which suffers from cross-layer covariate shift, we evaluate the frozen full-width attention and FFN operators directly on the student's hidden states $h_\ell$:
\begin{equation}
    \mathcal{L}_{\mathrm{branch}} = \sum_{\ell, q} \left\| F_{\ell,q}^{\mathrm{compact}}(h_\ell) - \operatorname{sg}\!\left[ F_{\ell,q}^{\mathrm{full}}(h_\ell) \right] \right\|_2^2,
    \label{eq:vae_branch}
\end{equation}
where $q$ indexes the attention and FFN branches.

Furthermore, since cache-free chunk decoding lacks persistent temporal states across chunk boundaries, independent chunk predictions can induce inter-frame flicker, particularly around fast-moving hands. Let $e_t = \widehat{I}_t - I_t$ denote the reconstruction error at frame $t$, and let $w_t^\Delta = 1 + \alpha_H (m_t \lor m_{t-1})$ upweight regions active in adjacent binary hand masks $m_t, m_{t-1}$. We enforce intra- and cross-chunk continuity:
\begin{equation}
    \mathcal{L}_{\Delta} = \langle w_t^\Delta |e_t - e_{t-1}| \rangle, \quad
    \mathcal{L}_{\partial} = \langle w_t^\Delta |e_t - e_{t-1}| \rangle_{t \in \mathcal{B}},
    \label{eq:vae_temporal}
\end{equation}
where $\mathcal{B}$ denotes chunk boundary frame indices. Ultra-Fast-VAE's end-to-end objective combines pixel reconstruction ($\mathcal{L}_{\mathrm{pix}}$), frozen teacher distillation ($\mathcal{L}_{\mathrm{teacher}}$), operator alignment ($\mathcal{L}_{\mathrm{branch}}$), temporal losses ($\mathcal{L}_{\Delta}, \mathcal{L}_{\partial}$), and LPIPS~\cite{lpips}/DINOv2~\cite{dinov2} perceptual losses:
\begin{equation}
    \mathcal{L}_{\mathrm{VAE}} = \lambda_p \mathcal{L}_{\mathrm{pix}} + \lambda_t \mathcal{L}_{\mathrm{teacher}} + \sum_{j\in\mathcal J} \lambda_j \mathcal{L}_j.
    \label{eq:vae_objective}
\end{equation}
Here $\lambda_p=50$, $\lambda_t=10$, and $\mathcal J$ indexes the 18 auxiliary entries in Appendix Table~\ref{tab:vae_losses}. Implementation details are in Appendix~\ref{sec:vae_training_details}; reconstruction metrics and timing protocols are in Appendix~\ref{sec:supp_vae_evaluation}.

\subsection{Decoder Architecture}
\label{sec:vae_architecture}
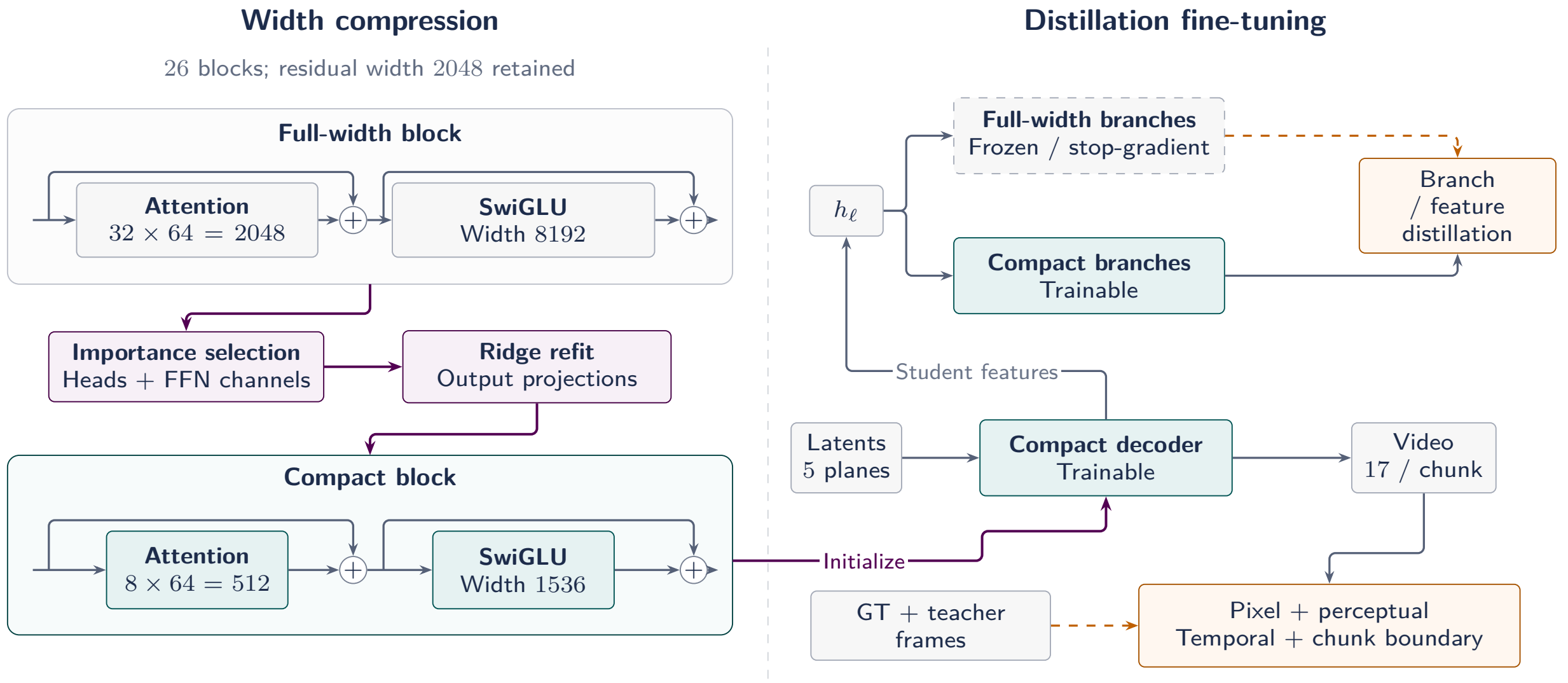
\begin{figure}[!htbp]
\centering
\resizebox{\textwidth}{!}{%
\begin{tikzpicture}[
  x=1cm,y=1cm,
  every node/.style={font=\sffamily\fontsize{8}{9.2}\selectfont,align=center,text=lynnfg},
  full/.style={draw=lynnfg!40,fill=gray!6,rounded corners=2pt,inner sep=3pt},
  compact/.style={draw=teal!65!black,fill=teal!10,rounded corners=2pt,inner sep=3pt},
  init/.style={draw=violet!55!black,fill=violet!6,rounded corners=2pt,inner sep=4pt,minimum height=.72cm},
  loss/.style={draw=orange!65!black,fill=orange!6,rounded corners=2pt,inner sep=4pt},
  sum/.style={draw=lynnfg!60,fill=white,circle,minimum size=.32cm,inner sep=0pt},
  flow/.style={-{Stealth[length=1.5mm]},draw=lynnfg!75,line width=.7pt,rounded corners=2pt},
  stop/.style={-{Stealth[length=1.5mm]},draw=orange!75!black,dashed,line width=.7pt,rounded corners=2pt},
  compress/.style={-{Stealth[length=1.7mm]},draw=violet!65!black,line width=.85pt,rounded corners=2pt}
]
\path[use as bounding box](-.15,.32) rectangle (18.65,8.63);
\node[font=\sffamily\bfseries\fontsize{10}{11}\selectfont] at (4.3,8.28)
  {Width compression};
\node[font=\sffamily\bfseries\fontsize{10}{11}\selectfont] at (13.85,8.28)
  {Distillation fine-tuning};
\draw[lynnfg!20,dashed] (9.02,.50)--(9.02,8.0);
\node[text=lynnfg!70,font=\sffamily\fontsize{8}{9.2}\selectfont] at (4.3,7.73)
  {$26$ blocks; residual width $2048$ retained};

\draw[draw=lynnfg!30,fill=gray!2,rounded corners=4pt] (0,5.18) rectangle (8.6,7.26);
\node[font=\sffamily\bfseries\fontsize{8.5}{10}\selectfont] at (4.3,6.98)
  {Full-width block};
\node[full,text width=2.65cm,minimum height=.88cm] (attnfull) at (2.25,5.94)
  {\textbf{Attention}\\$32\times64=2048$};
\node[sum] (addafull) at (4.10,5.94) {$+$};
\node[full,text width=2.90cm,minimum height=.88cm] (ffnfull) at (6.12,5.94)
  {\textbf{SwiGLU}\\Width $8192$};
\node[sum] (addffull) at (8.14,5.94) {$+$};
\draw[flow] (.30,5.94)--(attnfull.west);
\draw[flow] (attnfull.east)--(addafull.west);
\draw[flow] (addafull.east)--(ffnfull.west);
\draw[flow] (ffnfull.east)--(addffull.west);
\draw[flow] (addffull.east)--(8.43,5.94);
\draw[flow] (.49,5.94)--(.49,6.51)--(4.10,6.51)--(addafull.north);
\draw[flow] (4.46,5.94)--(4.46,6.51)--(8.14,6.51)--(addffull.north);

\node[init,text width=2.98cm] (select) at (2.12,4.20)
  {\textbf{Importance selection}\\Heads + FFN channels};
\node[init,text width=2.90cm] (refit) at (6.28,4.20)
  {\textbf{Ridge refit}\\Output projections};
\draw[compress] (4.30,5.18)--(4.30,4.83)--(2.12,4.83)--(select.north);
\draw[compress] (select.east)--(refit.west);
\draw[compress] (refit.south)--(6.28,3.40)--(4.30,3.40)--(4.30,3.15);

\draw[draw=teal!50!black,fill=teal!3,rounded corners=4pt] (0,1.02) rectangle (8.6,3.15);
\node[font=\sffamily\bfseries\fontsize{8.5}{10}\selectfont] at (4.3,2.87)
  {Compact block};
\node[compact,text width=1.94cm,minimum height=.92cm] (attnsmall) at (2.25,1.79)
  {\textbf{Attention}\\$8\times64=512$};
\node[sum] (addasmall) at (4.10,1.79) {$+$};
\node[compact,text width=1.94cm,minimum height=.92cm] (ffnsmall) at (6.12,1.79)
  {\textbf{SwiGLU}\\Width $1536$};
\node[sum] (addfsmall) at (8.14,1.79) {$+$};
\draw[flow] (.30,1.79)--(attnsmall.west);
\draw[flow] (attnsmall.east)--(addasmall.west);
\draw[flow] (addasmall.east)--(ffnsmall.west);
\draw[flow] (ffnsmall.east)--(addfsmall.west);
\draw[flow] (addfsmall.east)--(8.43,1.79);
\draw[flow] (.49,1.79)--(.49,2.38)--(4.10,2.38)--(addasmall.north);
\draw[flow] (4.46,1.79)--(4.46,2.38)--(8.14,2.38)--(addfsmall.north);

\node[full,text width=.66cm,minimum height=.60cm] (hidden) at (9.95,6.05) {$h_\ell$};
\node[full,dashed,text width=3.00cm,minimum height=.90cm] (teacher) at (12.83,6.94)
  {\textbf{Full-width branches}\\Frozen / stop-gradient};
\node[compact,text width=3.00cm,minimum height=.90cm] (branches) at (12.83,5.28)
  {\textbf{Compact branches}\\Trainable};
\node[loss,text width=2.05cm,minimum height=.98cm] (repr) at (17.20,6.11)
  {Branch / feature\\distillation};
\draw[flow] (hidden.east)--(10.66,6.05)--(10.66,6.94)--(teacher.west);
\draw[flow] (hidden.east)--(10.66,6.05)--(10.66,5.28)--(branches.west);
\draw[stop] (teacher.east)--(17.20,6.94)--(repr.north);
\draw[flow] (branches.east)--(17.20,5.28)--(repr.south);

\node[full,text width=1.10cm,minimum height=.83cm] (latents) at (9.95,3.12)
  {Latents\\$5$ planes};
\node[compact,text width=2.78cm,minimum height=.90cm] (decoder) at (13.03,3.12)
  {\textbf{Compact decoder}\\Trainable};
\node[full,text width=1.50cm,minimum height=.83cm] (frames) at (16.80,3.12)
  {Video\\$17$ / chunk};
\draw[flow] (latents.east)--(decoder.west);
\draw[flow] (decoder.east)--(frames.west);
\draw[compress] (8.60,1.90)--(11.56,1.90)--(11.56,2.25)--(13.03,2.25)--(decoder.south);
\node[text=violet!65!black,font=\sffamily\fontsize{7.5}{8.5}\selectfont,fill=white,inner sep=1pt] at (10.16,1.90)
  {Initialize};
\draw[flow] (decoder.north)--(13.03,4.15)--(9.95,4.15)--(hidden.south);
\node[text=lynnfg!65,font=\sffamily\fontsize{7.5}{8.5}\selectfont,fill=white,inner sep=1pt] at (11.50,4.15)
  {Student features};

\node[full,text width=2.63cm,minimum height=.82cm] (targets) at (10.95,1.13)
  {GT + teacher frames};
\node[loss,text width=4.24cm,minimum height=.98cm] (reconstruction) at (15.68,1.13)
  {Pixel + perceptual\\Temporal + chunk boundary};
\draw[stop] (targets.east)--(reconstruction.west);
\draw[flow] (frames.south)--(16.80,1.99)--(15.68,1.99)--(reconstruction.north);
\end{tikzpicture}
}
\caption{\textbf{Overview of Ultra-Fast-VAE compression and distillation.} \textbf{Left:} Structured width compression retains all 26 Transformer blocks and the 2048-dimensional residual stream while pruning intra-block attention heads and SwiGLU channels via activation-weighted norm scoring, followed by closed-form ridge regression initialization of output projections. \textbf{Right:} End-to-end fine-tuning aligns compact branches with frozen full-width operators on student hidden states $h_\ell$ alongside pixel, perceptual, and chunk-boundary temporal losses.}
\label{fig:vae_design}
\end{figure}

Figure~\ref{fig:vae_design} illustrates the structured width compression and distillation pipeline of Ultra-Fast-VAE. Built upon the H3-compatible LightVAE~\cite{lynnrealomni}, Ultra-Fast-VAE preserves all 26 Transformer blocks and the 2048-dimensional residual stream while pruning attention heads from 32 to 8 ($8\times64=512$ channels) and SwiGLU~\cite{swiglu} hidden width from 8192 to 1536, yielding 361.22M parameters. Following LightVAE's spatial tiling scheme, 28 overlapping $16\times16$ latent tiles are decoded and blended spatially. Each chunk processes 5 current latent planes (without future lookahead latents or persistent decoder KV caches) and outputs 17 RGB frames per chunk.

\subsection{Training and Quantized Export}
\label{sec:vae_training}
Training uses our 100K interaction corpus paired with native H3 latents; 1,394 videos are held out for validation and checkpoint selection. Each sample contains two consecutive 17-frame chunks, so supervision covers both within-chunk reconstruction and the transition between chunks. Pixel and perceptual targets preserve hand and object detail, while frozen full-width operators guide the compressed attention and feed-forward branches.

We average six late-stage checkpoints and export the decoder with SmoothQuant~\cite{smoothquant}, calibrated on training clips. W8A8 projections use per-token activation scales, per-channel weight scales, INT32 accumulation, and FP16 output. The loss definitions, weights, optimizer settings, and calibration protocol are given in Appendix~\ref{sec:vae_training_details}.

\subsection{Qualitative Reconstructions}
\label{sec:vae_reconstruction}
\begin{figure}[!htbp]
\centering
\includegraphics[width=.94\textwidth,height=.43\textheight,keepaspectratio]{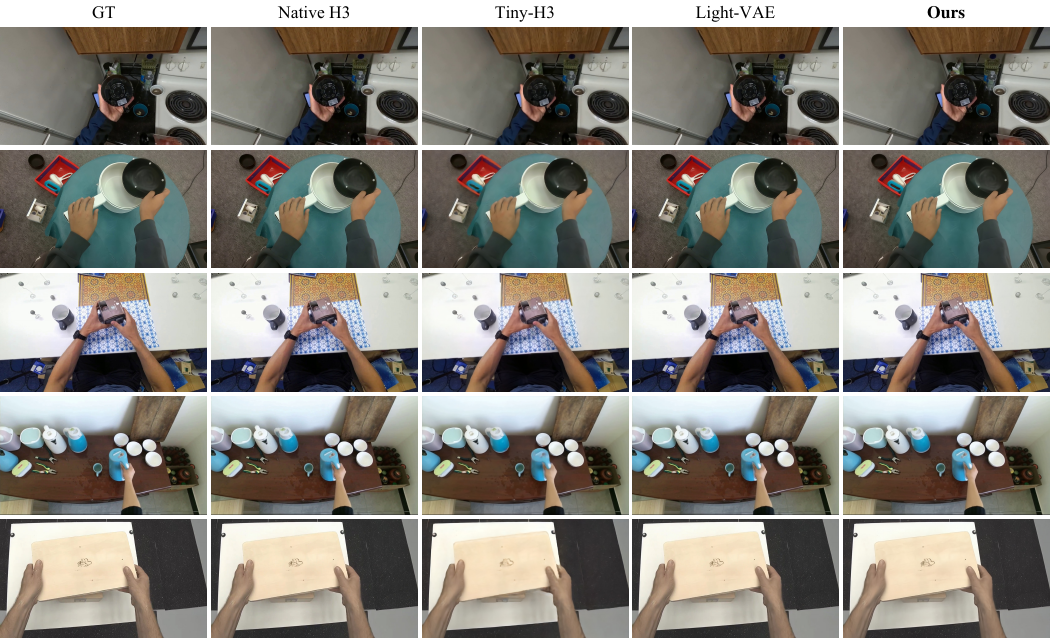}
\caption{\textbf{Full-frame VAE reconstruction comparisons across five interaction scenarios.} Each row compares outputs at an identical source timestamp across Native H3, Tiny-H3, LightVAE, and Ultra-Fast-VAE (Ours). While Tiny-H3 introduces noticeable blurring around articulated fingers and manipulated tools, Ultra-Fast-VAE faithfully recovers fine spatial textures comparable to full-width Transformer decoders.}
\label{fig:vae_reconstruction_objects}
\end{figure}

Figure~\ref{fig:vae_reconstruction_objects} compares full-frame reconstructions across five interaction scenarios, highlighting finger textures, tool edges, and background consistency across the decoder baselines.

\section{Data and HandBench}
\label{sec:report_data}

\subsection{Interaction Data and Aligned Annotations}
\label{sec:trainingdata}

To address the scarcity of large-scale annotated interaction data, we curate 100K video clips spanning real-world recordings and synthetic scenes across indoor, outdoor, and sci-fi environments. All clips are processed through a unified annotation pipeline to extract synchronized multimodal controls: a vision-language model generates scene and action captions, VIPE~\cite{vipe} estimates camera trajectories, and WiLoR~\cite{wilor} recovers per-frame MANO~\cite{mano} parameters. The reconstructed hand meshes are rasterized onto the image plane as color-coded conditioning videos (red for right hand, blue for left hand), ensuring strict spatiotemporal alignment with the target frames. Data sources and preprocessing details are provided in Appendix~\ref{sec:supp_data_engine}.

\subsection{HandBench: Interaction Benchmark}
\label{sec:benchmark}

\noindent\textbf{Benchmark Composition.}
We construct HandBench as a standardized testbed of 200 sequences (20 clips per source) covering bimanual manipulation, tool use, and interactions with articulated or deformable objects. Each sequence preserves its original temporal interval, spatial crop, and aligned hand and camera conditions. The benchmark is partitioned into a 160-clip in-domain main split sampled across eight training sources (EgoDex~\cite{egodex}, HOI-4D~\cite{hoi4d}, OpenAoE~\cite{openaoe}, ARCTIC~\cite{arctic}, CaptainCook4D~\cite{captaincook4d}, Ego4D~\cite{ego4d}, EPIC-KITCHENS-100~\cite{epickitchens100}, and synthetic scenes from H3~\cite{minimaxh3}) and a 40-clip out-of-distribution (OOD) split comprising H2O~\cite{h2o} and TACO~\cite{taco}, which are held out entirely from training to evaluate cross-dataset generalization.

\noindent\textbf{Evaluation Metrics and Protocol.}
We evaluate generated videos across four dimensions. Hand control accuracy is measured by the silhouette Dice score between MANO meshes reconstructed from generated frames and ground-truth masks. Visual and distribution fidelity are evaluated via PSNR, SSIM~\cite{ssim}, LPIPS~\cite{lpips}, and FVD~\cite{fvd}. We assess semantic alignment via weighted CLIP~\cite{clip} similarity, temporal coherence via VBench~\cite{vbench} subject and background consistency, and camera adherence via background motion agreement with reference videos. All models are evaluated at their native sampling rates over a shared spatiotemporal support, averaging scalar metrics equally across sources and pooling features for FVD (definitions: Appendix~\ref{sec:supp_metrics}).

\begin{table}[!t]
\centering\scriptsize
\setlength{\tabcolsep}{3pt}
\renewcommand{\arraystretch}{1.02}
\caption{\textbf{HandBench results.} Top: eight-source in-domain split; bottom: held-out H2O and TACO. \textbf{Bold}/\underline{underlining}: best/second-best per cohort. FT: fine-tuning on our 100K corpus; SC/BC: VBench subject/background consistency. Metrics and protocols: Appendices~\ref{sec:supp_metrics} and~\ref{sec:supp_timing}.}
\label{tab:HandBench}
\begin{adjustbox}{width=\linewidth}
\begin{tabular}{lrrrrrrrrr@{}}
\toprule
Method & FVD $\downarrow$ & PSNR $\uparrow$ & LPIPS $\downarrow$ & SSIM $\uparrow$ & SC $\uparrow$ & BC $\uparrow$ & \shortstack{Weighted\\CLIP $\uparrow$} & \shortstack{MANO\\Dice $\uparrow$} & Cam-BG $\uparrow$ \\
\midrule
JointControlVideo~\cite{jointcontrolvideo} & 358.34 & 16.68 & 0.3460 & 0.6904 & 93.77 & 94.34 & 82.15 & 89.93 & 23.21 \\
Generated Reality~\cite{generatedreality} & 348.26 & 17.03 & 0.3358 & 0.7022 & 94.56 & 94.33 & 82.03 & 89.11 & 26.65 \\
Hand2World (bidir., FT)~\cite{hand2world} & 294.32 & \underline{19.64} & \underline{0.2462} & \underline{0.7460} & 94.05 & 94.22 & 81.96 & 90.60 & \underline{54.13} \\
Hand2World (streaming)~\cite{hand2world} & 963.96 & 12.92 & 0.5554 & 0.6594 & 80.96 & 88.99 & 73.44 & 76.23 & 17.17 \\
\textbf{LynnReal-Hand-Base} & \textbf{181.67} & \textbf{20.90} & \textbf{0.2071} & \textbf{0.7605} & \underline{94.94} & \underline{94.77} & \textbf{83.59} & \textbf{93.02} & \textbf{64.97} \\
\textbf{LynnReal-Hand-Fast} & \underline{237.33} & 19.12 & 0.2617 & 0.7402 & \textbf{95.27} & \textbf{95.34} & \underline{82.85} & \underline{92.49} & 49.35 \\
\bottomrule
\end{tabular}
\end{adjustbox}
\par\smallskip
\setlength{\tabcolsep}{2pt}
\begin{adjustbox}{width=\linewidth}
\begin{tabular}{lrrrrrrrrrr@{}}
\textit{OOD evaluation} & \multicolumn{5}{c}{H2O} & \multicolumn{5}{c}{TACO} \\
\cmidrule(lr){2-6}\cmidrule(lr){7-11}
Method & FVD $\downarrow$ & PSNR $\uparrow$ & LPIPS $\downarrow$ & \shortstack{MANO\\Dice $\uparrow$} & Cam-BG $\uparrow$ & FVD $\downarrow$ & PSNR $\uparrow$ & LPIPS $\downarrow$ & \shortstack{MANO\\Dice $\uparrow$} & Cam-BG $\uparrow$ \\
\midrule
JointControlVideo~\cite{jointcontrolvideo} & 1038.22 & 14.77 & 0.3691 & 90.75 & 16.69 & 942.40 & 17.02 & 0.4040 & 89.96 & 17.97 \\
Generated Reality~\cite{generatedreality} & 1146.33 & 16.24 & 0.3177 & 87.90 & 21.49 & 1019.89 & 16.88 & 0.4230 & 88.31 & 25.25 \\
Hand2World (bidir., FT)~\cite{hand2world} & 879.73 & \underline{18.67} & 0.2275 & 92.25 & \underline{56.10} & 762.02 & 17.94 & \underline{0.3630} & \underline{92.30} & \underline{64.34} \\
Hand2World (streaming)~\cite{hand2world} & 3031.95 & 12.71 & 0.5718 & 80.01 & 20.85 & 2800.99 & 12.87 & 0.6603 & 70.46 & 12.72 \\
\textbf{LynnReal-Hand-Base} & \textbf{434.46} & \textbf{20.32} & \textbf{0.1709} & \textbf{93.70} & \textbf{64.75} & \textbf{398.03} & \textbf{19.59} & \textbf{0.2746} & \textbf{92.84} & \textbf{68.99} \\
\textbf{LynnReal-Hand-Fast} & \underline{690.19} & 18.61 & \underline{0.2249} & \underline{92.78} & 51.22 & \underline{630.39} & \underline{18.08} & 0.3644 & 91.67 & 54.24 \\
\bottomrule
\end{tabular}
\end{adjustbox}
\end{table}

\section{Experimental Results}
\label{sec:experiments}

\textbf{Implementation and baseline protocols.}
We fine-tune bidirectional Hand2World~\cite{hand2world} and Generated Reality~\cite{generatedreality} on our 100K training set, while evaluating released checkpoints for JointControlVideo~\cite{jointcontrolvideo} and streaming Hand2World due to prohibitive training costs and unavailable distillation code, respectively. Unless noted, \texttt{LynnReal-Hand-Base} and all baselines generate at 768p, whereas \texttt{LynnReal-Hand-Fast} runs at 540p for real-time streaming; all predictions are evaluated against 768p ground-truth references (Appendices~\ref{sec:supp_metrics} and~\ref{sec:supp_inference_variants}).

\subsection{Quantitative Results}
\label{sec:quantitative_results}

\textbf{In-domain and OOD interaction quality.}
As shown in Table~\ref{tab:HandBench} (top), \texttt{LynnReal-Hand-Base} achieves state-of-the-art performance across all visual, consistency, and control metrics on the eight-source benchmark. Compared to the strongest fine-tuned baseline, Hand2World (bidir., FT), our Base model reduces FVD by 38.3\% (181.67 vs.\ 294.32) while improving MANO Dice from 90.60 to 93.02 and camera adherence (Cam-BG) from 54.13 to 64.97. Without adaptation to our 100K corpus, the released Hand2World (streaming) checkpoint degrades sharply on our diverse benchmark (963.96 FVD and 76.23 MANO Dice). Our three-step \texttt{LynnReal-Hand-Fast} retains a 92.49 MANO Dice score and achieves the highest subject and background consistency (95.27 SC / 95.34 BC), outperforming all offline bidirectional baselines in FVD (237.33). This holds on unseen H2O and TACO OOD splits (Table~\ref{tab:HandBench}, bottom), where Base cuts FVD by over 47\% relative to Hand2World (bidir., FT) and Fast generalizes without retraining. Fast retains hand alignment within 0.53 Dice points of Base, while its lower Cam-BG (49.35 vs.\ 64.97) reveals a camera-following trade-off.

\textbf{End-to-end streaming and decoding efficiency.}
Tables~\ref{tab:generation_efficiency} and~\ref{tab:decoder} evaluate generation and decoding throughput on a single H200 GPU. By combining three-step AnyFlow transitions with Ultra-Fast-VAE, \texttt{LynnReal-Hand-Fast} synthesizes 17-frame chunks at 540p with 0.741\,s latency, 18.03\,FPS throughput, and 34.89\,GiB peak memory. In standalone 768p decoding benchmarks (Table~\ref{tab:decoder}), Ultra-Fast-VAE reduces parameter count by $6.7\times$ relative to Native H3 (361.2M vs.\ 2423.5M) and accelerates 17-frame decoding by $17.8\times$ (46.77\,ms vs.\ 831.30\,ms), achieving 34.43\,dB PSNR. Compared with Tiny-H3, Ultra-Fast-VAE improves PSNR by 4.03\,dB (34.43 vs.\ 30.40) and reduces temporal error from 13.78 to 9.80. It matches StreamRender~\cite{streamrender} in perceptual and temporal quality at $2.93$--$3.06\times$ the throughput. Relative to LightVAE, 0.70\,dB lower PSNR accompanies $12.5\times$ lower latency, balancing reconstruction fidelity and streaming efficiency.

\begin{table}[!t]
\centering\scriptsize
\setlength{\tabcolsep}{2pt}
\renewcommand{\arraystretch}{1.02}
\caption{\textbf{End-to-end generation efficiency on a single H200 GPU.} Deployment settings are given in Appendix~\ref{sec:supp_fast_timing}.}
\label{tab:generation_efficiency}
\begin{adjustbox}{max width=\linewidth}
\begin{tabular}{@{}lrrrrr@{}}
\toprule
Method & Params & Timed unit & \shortstack{Time\\(s) $\downarrow$} & FPS $\uparrow$ & \shortstack{Peak\\(GiB)} \\
\midrule
JointControlVideo~\cite{jointcontrolvideo} & 14B & Full seq. & 1320.324 & 0.053 & 59.50 \\
Generated Reality~\cite{generatedreality} & 5B & Full seq. & 446.484 & 0.221 & 15.87 \\
Hand2World (bidir., FT)~\cite{hand2world} & 5B & Full seq. & 104.682 & 1.195 & 59.67 \\
Hand2World (streaming)~\cite{hand2world} & 5B & 4 frames & 0.463 & 8.646 & 33.44 \\
\textbf{LynnReal-Hand-Base} & 33B & Full seq. & 770.155 & 0.136 & 82.59 \\
\textbf{LynnReal-Hand-Fast} & 33B & 17 frames & 0.741 & 18.03 & 34.89 \\
\bottomrule
\end{tabular}
\end{adjustbox}
\end{table}

\begin{table}[!t]
\centering
\captionsetup{font=footnotesize,skip=4pt}
\caption{\textbf{Decoder reconstruction quality and latency at $1344\times768$.} Latency and FPS are measured per 17-frame chunk on an H200 GPU. StreamRender\textsuperscript{1/2} decodes 1/2 latents per step, LightVAE~\cite{lynnrealomni} uses 2 lookahead latents, and Ultra-Fast-VAE runs cache-free with W8A8 quantization. Temporal error denotes frame-difference MAE ($\times 10^3$).}
\label{tab:decoder}
\footnotesize
\setlength{\tabcolsep}{1.1pt}
\renewcommand{\arraystretch}{1.05}
\begin{adjustbox}{max width=\linewidth}
\begin{tabular}{lrrrrrr@{}}
\toprule
Decoder & \shortstack{Params\\(M) $\downarrow$} & \shortstack{PSNR\\(dB) $\uparrow$} & \shortstack{LPIPS $\downarrow$} & \shortstack{Temporal\\error $\downarrow$} & \shortstack{Latency\\(ms) $\downarrow$} & \shortstack{FPS\\$\uparrow$} \\
\midrule
Native H3~\cite{minimaxh3} & 2423.5 & \underline{35.04} & \underline{0.0434} & \underline{9.05} & 831.30 & 20.44 \\
Tiny-H3 & \textbf{9.9} & 30.40 & 0.1040 & 13.78 & \textbf{10.90} & \textbf{1555.64} \\
StreamRender\textsuperscript{1}~\cite{streamrender} & 1617.8 & 33.74 & 0.0533 & 10.09 & 143.31 & 118.63 \\
StreamRender\textsuperscript{2}~\cite{streamrender} & 1617.8 & 33.95 & 0.0516 & 9.88 & 157.65 & 123.91 \\
SolarX V3~\cite{solarxvae} & 2423.5 & 34.09 & 0.0484 & 10.20 & 552.79 & 30.54 \\
LightVAE~\cite{lynnrealomni} & 1752.1 & \textbf{35.13} & \textbf{0.0423} & \textbf{8.97} & 586.40 & 28.97 \\
\midrule
\textbf{Ultra-Fast-VAE} & \underline{361.2} & 34.43 & 0.0511 & 9.80 & \underline{46.77} & \underline{363.21} \\
\bottomrule
\end{tabular}
\end{adjustbox}
\end{table}

\begin{table}[!htbp]
\centering\footnotesize
\captionsetup{skip=3pt}
\setlength{\tabcolsep}{5pt}
\renewcommand{\arraystretch}{0.96}
\caption{\textbf{Pairwise human preference (\%).} Win rates of LynnReal-Hand-Base/Fast against baselines, following~\cite{selfforcing}.}
\label{tab:user_preference}
\begin{adjustbox}{max width=\linewidth}
\begin{tabular}{lcc@{}}
\toprule
Baseline & LynnReal-Hand-Base $\uparrow$ & LynnReal-Hand-Fast $\uparrow$ \\
\midrule
JointControlVideo~\cite{jointcontrolvideo} & 93.50 & 88.00 \\
Generated Reality~\cite{generatedreality} & 89.00 & 84.00 \\
Hand2World (bidir., FT)~\cite{hand2world} & 70.50 & 58.50 \\
Hand2World (streaming)~\cite{hand2world} & 100.00 & 100.00 \\
\bottomrule
\end{tabular}
\end{adjustbox}
\end{table}
\textbf{Human preference study.}
Following~\cite{selfforcing}, we conduct pairwise blind user evaluations across motion accuracy and visual realism (Table~\ref{tab:user_preference}). Evaluators consistently favor both \texttt{LynnReal-Hand-Base} and \texttt{LynnReal-Hand-Fast} over all competing methods, achieving 70.5\% and 58.5\% win rates even against the offline bidirectional Hand2World (bidir., FT), and 100\% preference over its streaming counterpart.

\subsection{Qualitative Results}
\label{sec:qualitative_results}
\begin{figure}[!htbp]
\centering
\includegraphics[width=\textwidth]{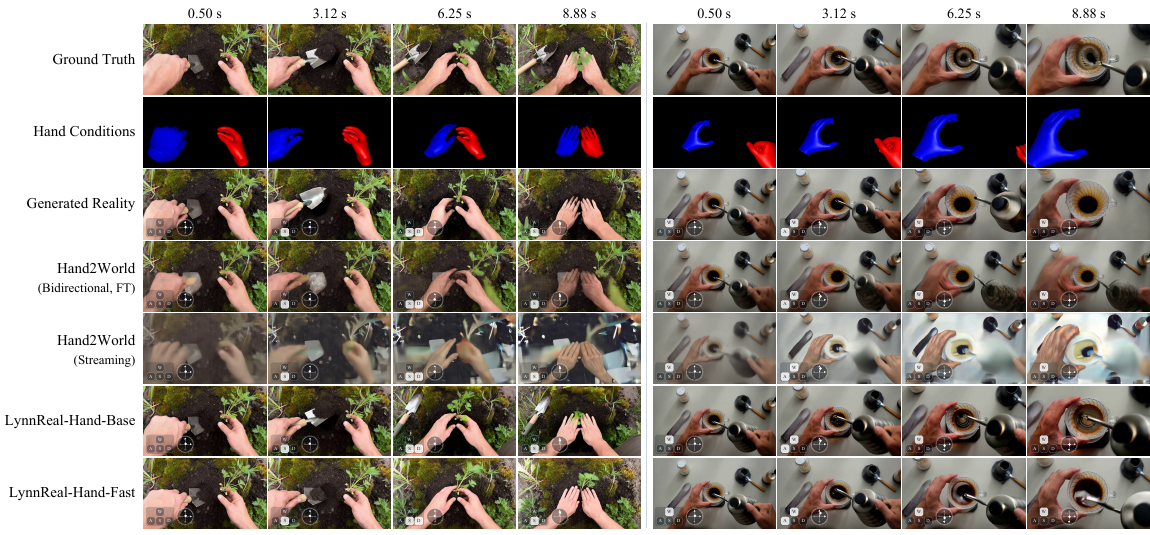}
\caption{\textbf{Hand- and camera-conditioned generation.} Four source-aligned frames per sequence are shown with GT and camera-trajectory overlays. Base and Fast denote LynnReal-Hand-Base and LynnReal-Hand-Fast, respectively. Both preserve finger articulation and object interaction under camera motion. More in Sec.~\ref{sec:qualitative_comparisons}.}
\label{fig:demo_qualitative}
\end{figure}

Figure~\ref{fig:demo_qualitative} presents visual comparisons across complex hand-object interactions under prescribed hand and camera trajectories. Competing baselines frequently exhibit merged fingers, severe motion blur during rapid manipulation, or background warping when camera viewpoints shift. Both \texttt{LynnReal-Hand-Base} and \texttt{LynnReal-Hand-Fast} maintain crisp finger-object contact boundaries and coherent scene geometry across time. Figure~\ref{fig:vae_reconstruction_objects} compares the VAE reconstructions discussed in Sec.~\ref{sec:vae_reconstruction}. Tiny-H3 smooths skin textures and tool edges, whereas Ultra-Fast-VAE preserves finger contours and object boundaries comparable to Native H3 and LightVAE. These visual comparisons complement the reconstruction metrics: hand regions occupy few pixels, so visually consequential errors around fingers and contact boundaries can be less apparent in aggregate reconstruction scores.

\begin{figure}[!htbp]
\centering
\includegraphics[width=\textwidth,height=.88\textheight,keepaspectratio]{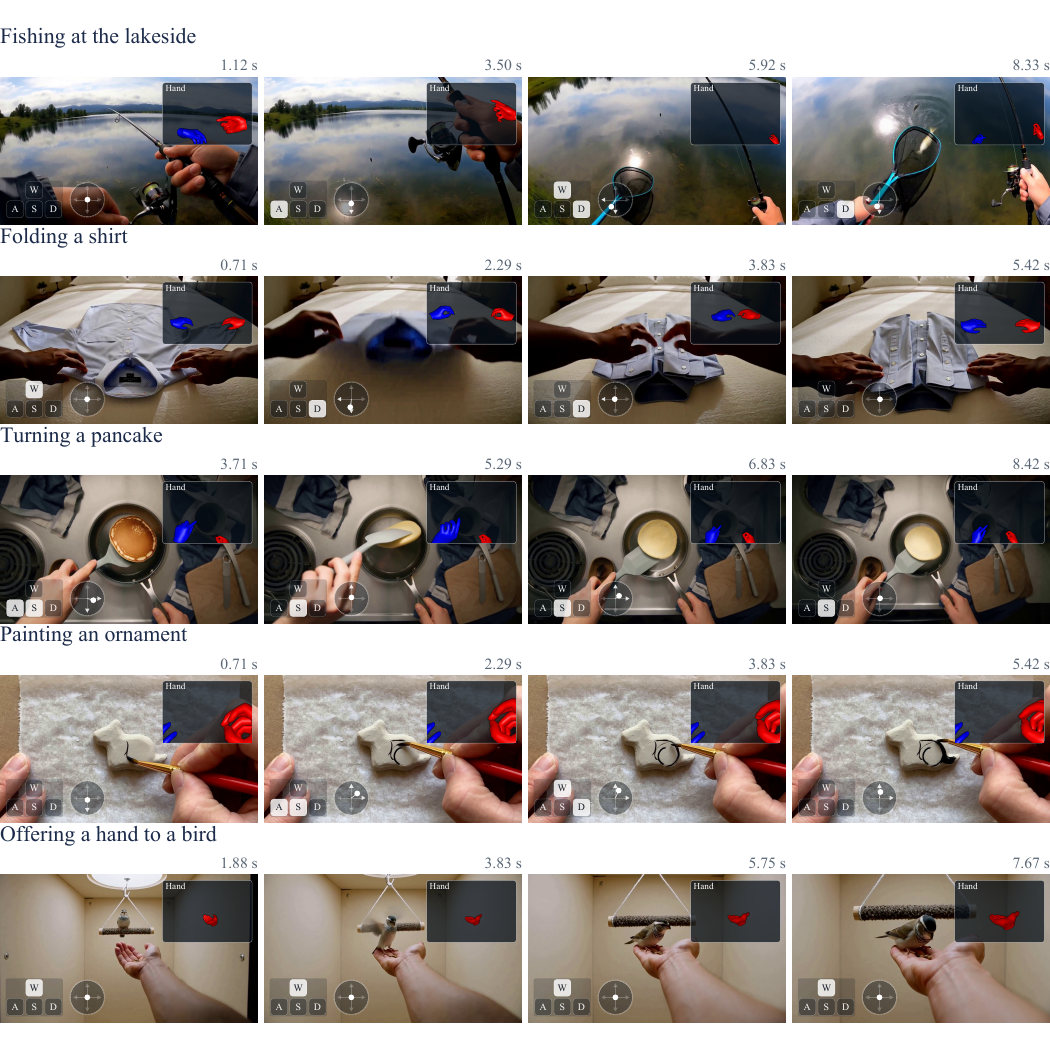}
\caption{Hand-conditioned interaction in everyday scenes. Each row follows one demo sequence. Upper-right insets show the aligned MANO hand conditions within the generated LynnReal-Hand-Base frames. Overlaid WASD and rotation controls visualize the supplied camera motion, following Figure~\ref{fig:demo_qualitative}; time progresses left to right. The strips cover fishing, cloth folding, cooking, ornament painting, and bird interaction.}
\label{fig:showcase_everyday}
\end{figure}

\begin{figure}[!htbp]
\centering
\includegraphics[width=\textwidth,height=.82\textheight,keepaspectratio]{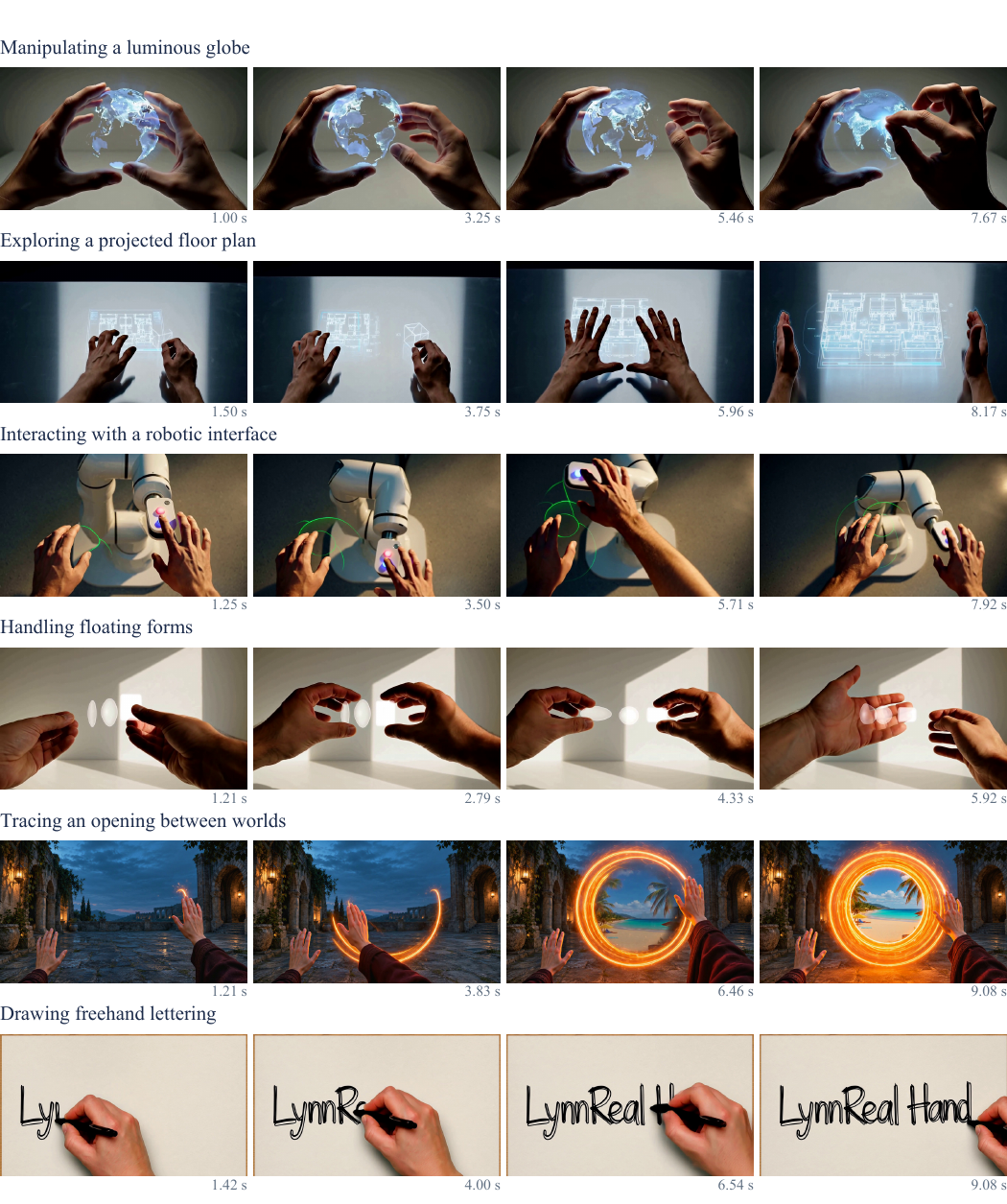}
\caption{Selected imaginative interactions generated by LynnReal-Hand-Base. Each row follows one saved demo video, from manipulating luminous forms and projected layouts to tracing a portal and writing on paper. Only generated outputs are shown; frame times refer to the original videos.}
\label{fig:showcase_imaginative}
\end{figure}

\paragraph{Additional interaction demonstrations.}
Figures~\ref{fig:showcase_everyday} and~\ref{fig:showcase_imaginative} show additional LynnReal-Hand-Base demonstrations in everyday and imaginative scenes. The everyday examples include aligned hand-condition insets and supplied camera-motion overlays; the imaginative examples present the generated sequences.

\subsubsection{Additional Qualitative Comparisons}
\label{sec:qualitative_comparisons}

\paragraph{Visualization Protocol.}
\label{sec:qualitative_protocol}
All qualitative comparisons display full native viewports scaled uniformly with aspect-preserving padding at aligned source timestamps, without post-processing or temporal interpolation. Model configurations match Table~\ref{tab:HandBench}.

\paragraph{Conditioned Interaction Demonstrations.}
\label{sec:interaction_demonstrations}
Figure~\ref{fig:demo_qualitative} compares methods on synthetic seedling-transplanting and pour-over coffee sequences at four aligned timestamps ($0.50$, $3.125$, $6.25$, and $8.875$\,s).

\begin{figure}[!htbp]
\centering
\includegraphics[width=\textwidth,height=.39\textheight,keepaspectratio]{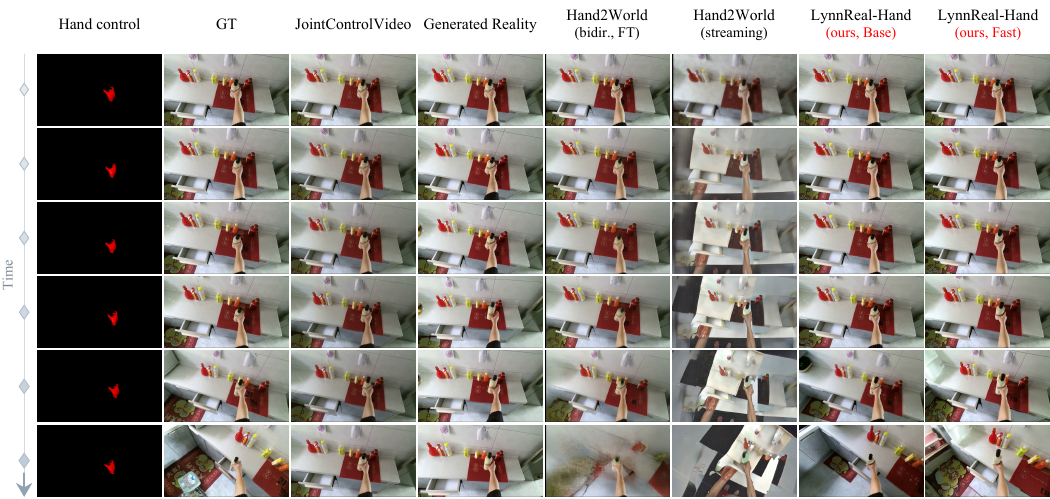}
\caption{\textbf{Qualitative comparison on bottle manipulation under viewpoint changes (HOI-4D).} Six source-aligned frames are shown across input hand conditions, ground truth, and the six evaluated models in Table~\ref{tab:HandBench}, demonstrating stable object geometry and grasp alignment during camera movement.}
\label{fig:handbench_more_hoi4d}
\end{figure}

\paragraph{Additional HandBench Comparison.}
\label{sec:handbench_visual_comparison}
Figure~\ref{fig:handbench_more_hoi4d} provides an additional HandBench comparison on bottle manipulation at six uniformly spaced timestamps, illustrating hand-motion adherence and object stability during camera movement.

\begin{figure}[!htbp]
\centering
\includegraphics[width=\textwidth]{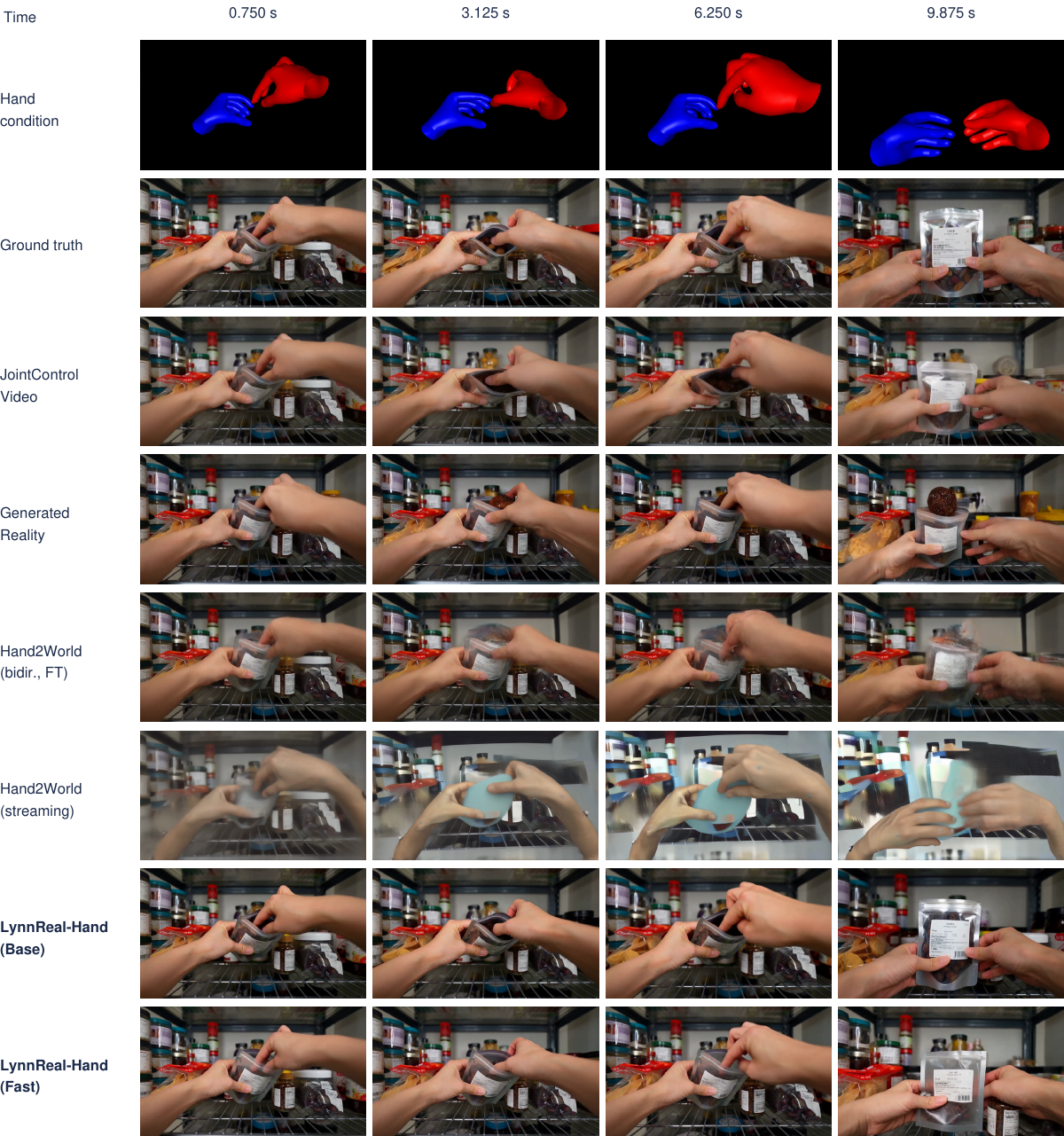}
\caption{\textbf{Bimanual container handling (HB0557).} Rows show hand conditions, ground truth, and the six models in Table~\ref{tab:HandBench}; columns use the same four source timestamps for every method. The sequence tests hand articulation and container geometry during manipulation inside a refrigerator.}
\label{fig:synthetic_hb0557}
\end{figure}

\begin{figure}[!htbp]
\centering
\includegraphics[width=\textwidth]{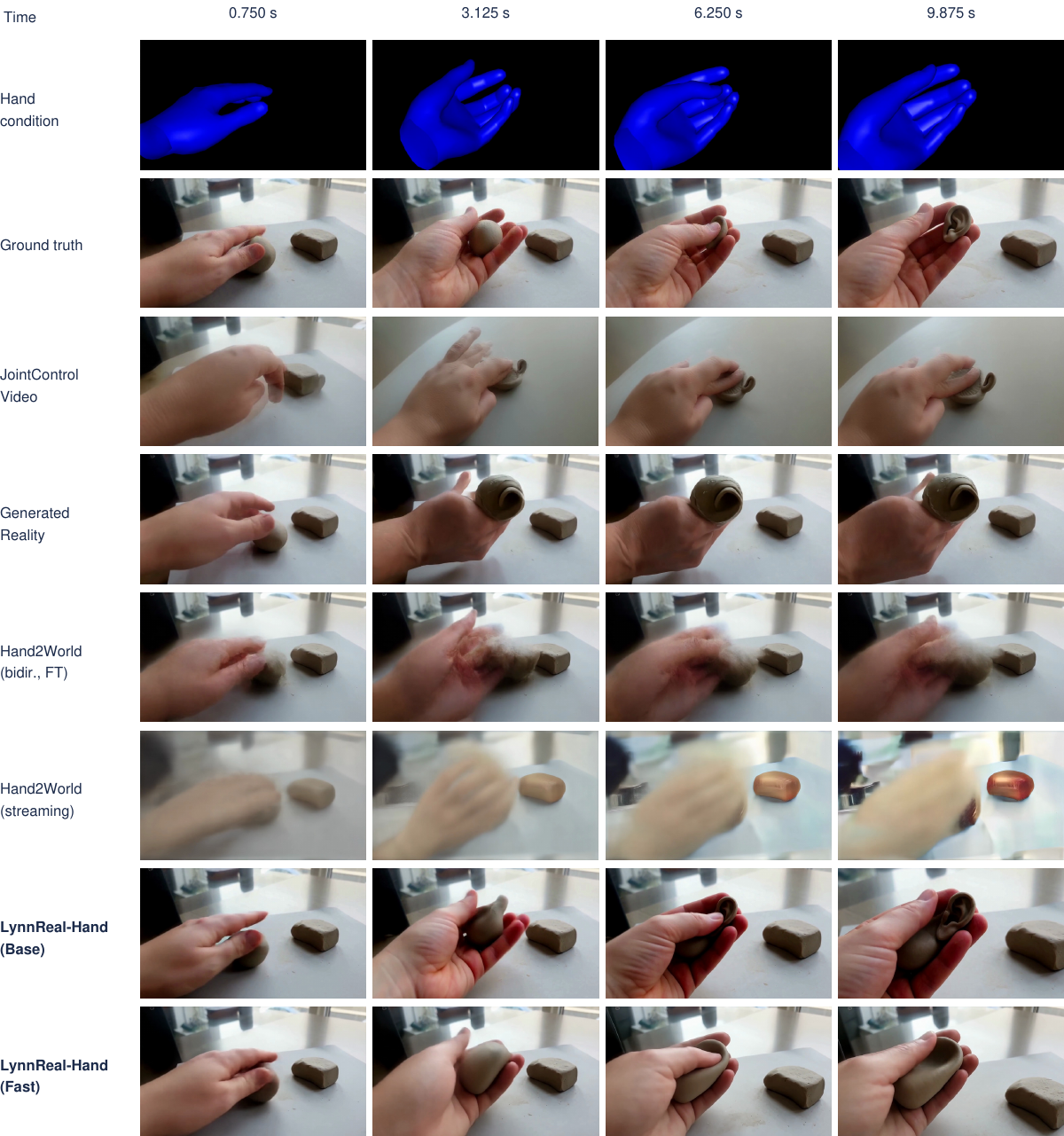}
\caption{\textbf{Hand-held object shaping (HB0559).} Method rows and source-timestamp columns follow Fig.~\ref{fig:synthetic_hb0557}. This case shows finger contact and changes in a small object's shape.}
\label{fig:synthetic_hb0559}
\end{figure}

\begin{figure}[!htbp]
\centering
\includegraphics[width=\textwidth]{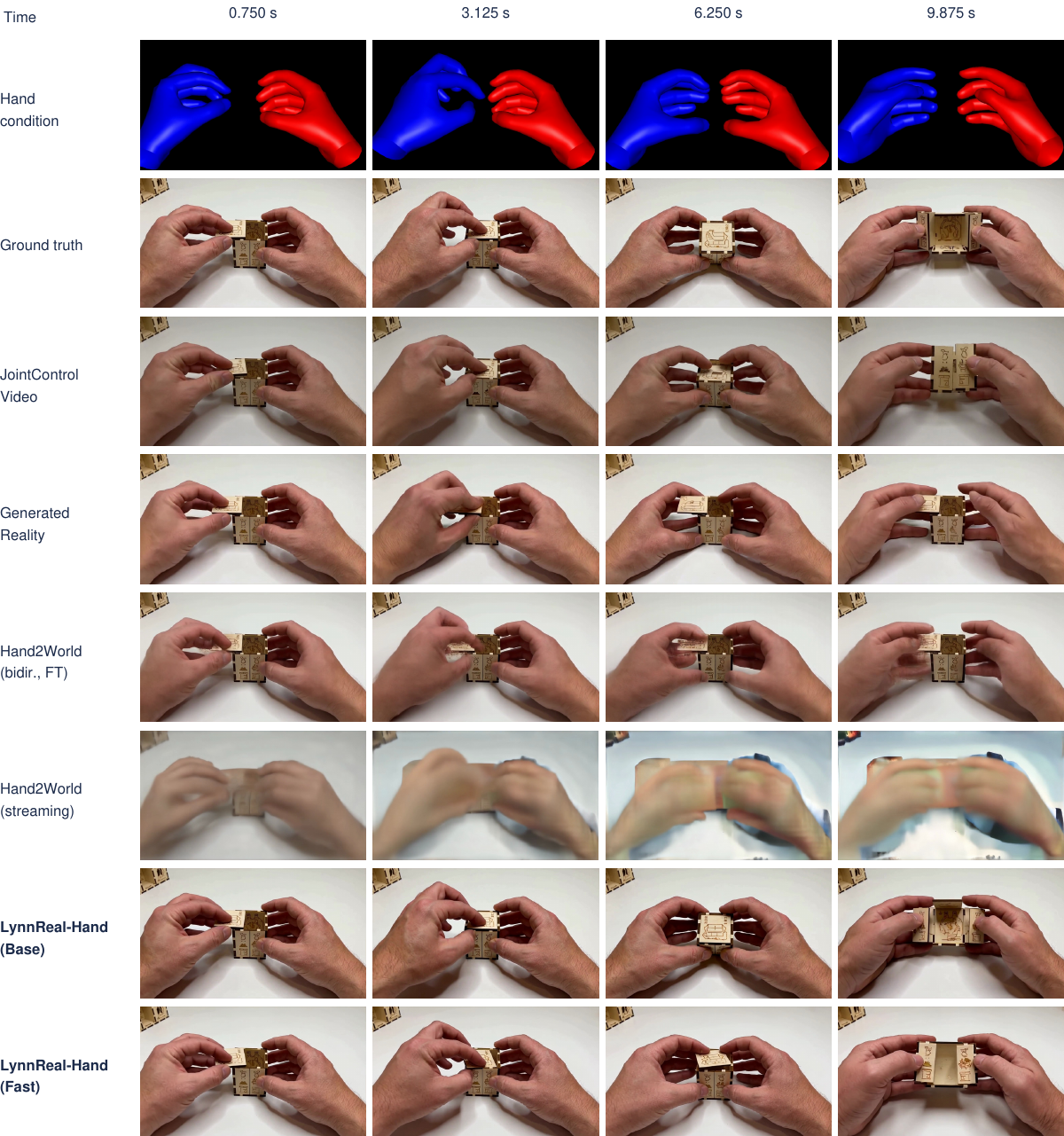}
\caption{\textbf{Bimanual wooden-puzzle manipulation (HB0562).} Method rows and source-timestamp columns follow Fig.~\ref{fig:synthetic_hb0557}. Both hands manipulate a small patterned object, exposing errors in finger alignment and object structure.}
\label{fig:synthetic_hb0562}
\end{figure}

\begin{figure}[!htbp]
\centering
\includegraphics[width=\textwidth]{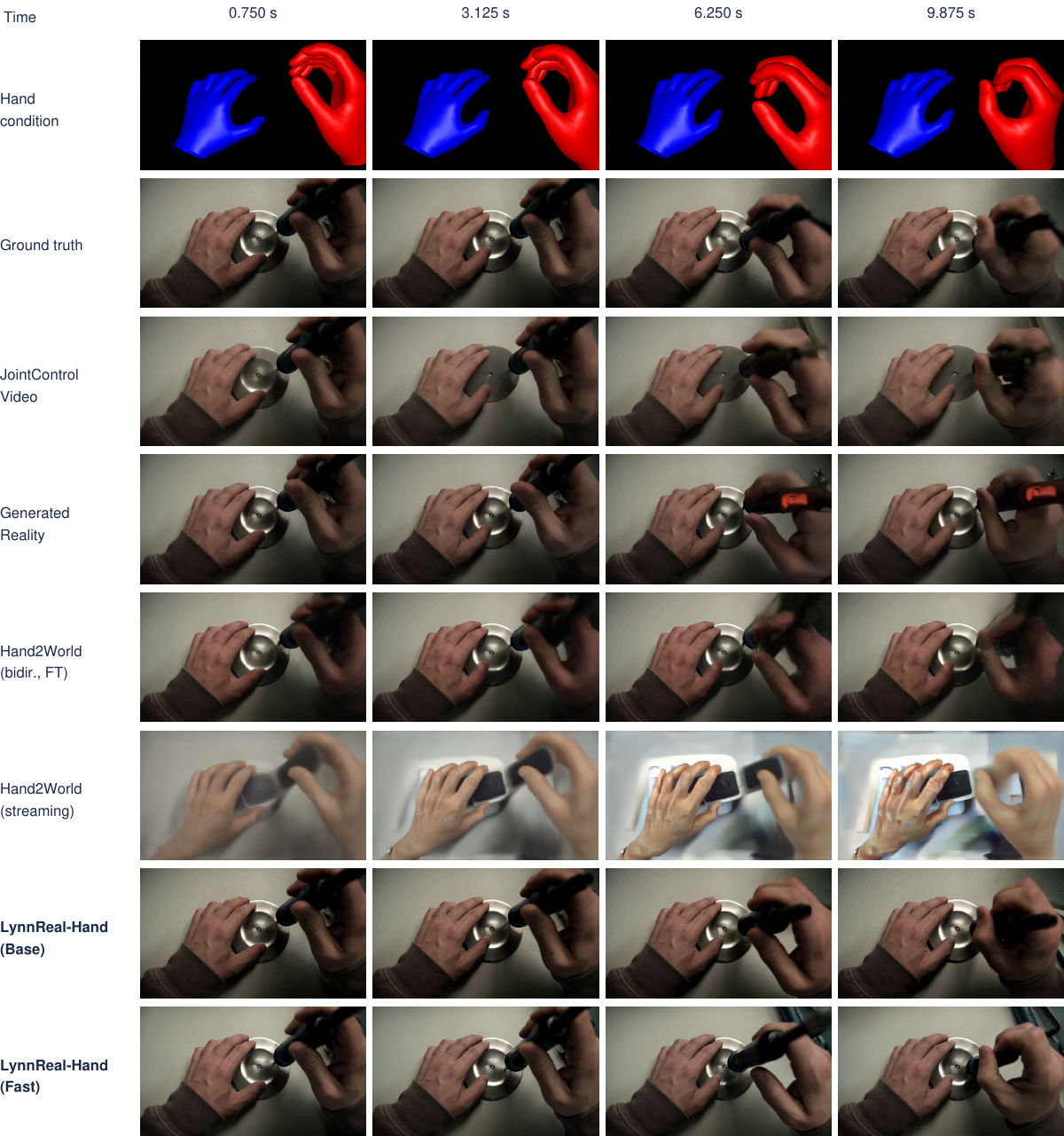}
\caption{\textbf{Bimanual tool interaction (HB0565).} Method rows and source-timestamp columns follow Fig.~\ref{fig:synthetic_hb0557}. One hand stabilizes a metal object while the other positions a tool against it.}
\label{fig:synthetic_hb0565}
\end{figure}

\paragraph{Additional Synthetic Interaction Cases.}
Figures~\ref{fig:synthetic_hb0557}--\ref{fig:synthetic_hb0565} compare the six evaluated models on four additional H3-synthetic cases: container handling, small-object shaping, wooden-puzzle manipulation, and tool use. Each comparison includes the hand conditions and ground truth at four shared source timestamps. The full image panels are preserved and scaled uniformly for display.

\FloatBarrier
\subsection{Ablation Study}
\label{sec:ablation_results}

\textbf{Multi-level hand supervision (Stage~I).}
Table~\ref{tab:stage1_ablation} isolates the contribution of each hand-aware loss component in Stage~I. To limit full-sequence inference cost, all variants use the same 100 cases randomly selected from HandBench's 200 cases; this evaluation cohort differs from the main and OOD splits in Table~\ref{tab:HandBench}. Training with global flow matching alone yields suboptimal hand alignment (87.02 MANO Dice) and high FVD (473.69). Adding latent ROI supervision ($\mathcal{L}_{\mathrm{ROI}}$) improves hand geometry by +2.67 Dice points, while introducing decoded perceptual supervision ($\mathcal{L}_{\mathrm{LPIPS}}^H$) produces the largest leap in visual realism and articulation accuracy, reducing FVD from 442.33 to 358.62 and boosting Dice to 92.43. Finally, incorporating semantic patch alignment ($\mathcal{L}_{\mathrm{DINO}}^H$) achieves the best overall performance (351.55 FVD, 92.74 Dice), confirming complementary latent and pixel-level supervision for fine hand control. The larger gain from perceptual supervision suggests that latent reconstruction leaves image-space errors, while semantic alignment adds constraints on visible hand structure.

\begin{table}[!htbp]
\centering\scriptsize
\setlength{\tabcolsep}{1.5pt}
\renewcommand{\arraystretch}{1.05}
\caption{\textbf{Stage-I hand-supervision ablation.} All variants use the same 100 cases randomly selected from the 200-case HandBench to limit full-sequence inference cost. ROI, LPIPS, and DINO denote latent, perceptual, and semantic hand losses. \textbf{Bold}/\underline{underlining}: best/second-best.}
\label{tab:stage1_ablation}
\begin{adjustbox}{max width=\linewidth}
\begin{tabular}{cccrrrrrr@{}}
\toprule
\multicolumn{3}{c}{Loss} & \multicolumn{6}{c}{Metrics} \\
\cmidrule(lr){1-3}\cmidrule(lr){4-9}
ROI & \shortstack{LPIPS} & \shortstack{DINO} & \shortstack{FVD $\downarrow$} & PSNR $\uparrow$ & \shortstack{LPIPS $\downarrow$} & SC $\uparrow$ & \shortstack{MANO\\Dice $\uparrow$} & Cam-BG $\uparrow$ \\
\midrule
$\times$ & $\times$ & $\times$ & 473.69 & 19.52 & 0.2471 & 94.28 & 87.02 & 57.64 \\
$\surd$ & $\times$ & $\times$ & 442.33 & 19.56 & 0.2463 & 94.36 & 89.69 & 61.26 \\
$\surd$ & $\surd$ & $\times$ & \underline{358.62} & \underline{20.79} & \underline{0.2158} & \underline{94.49} & \underline{92.43} & \underline{64.91} \\
$\surd$ & $\surd$ & $\surd$ & \textbf{351.55} & \textbf{21.21} & \textbf{0.2057} & \textbf{94.85} & \textbf{92.74} & \textbf{64.94} \\
\bottomrule
\end{tabular}
\end{adjustbox}
\end{table}

\textbf{Region-aware causal refinement (Stage~III).}
Table~\ref{tab:stage3_ablation} ablates Stage~III components on a 50-case \emph{full-hand} subset selected from HandBench, with hands visible in every frame. All variants use the same cases and three-step causal budget for a fair comparison. Directly rolling out the Stage~II flow-map initializer incurs exposure bias, degrading temporal quality (404.68 FVD) and hand accuracy (89.60 Dice). Adding video-score distillation under a standard noise schedule ($\text{shift}=1$) reduces drift (303.61 FVD), while shifting video evaluation toward higher noise ($\text{shift}=12$) strengthens kinematic alignment, raising MANO Dice from 90.53 to 92.03 and improving FVD (289.05) and PSNR (18.02). Incorporating an unmasked image-score branch ($\checkmark$) sharpens spatial fidelity (208.75 FVD, 18.89 PSNR) but stalls hand accuracy (92.00 Dice) due to conflicting gradients from the hand-blind image teacher. Restricting image supervision to background regions ($\checkmark(\text{BG})$) resolves this conflict, advancing MANO Dice to 92.50 while achieving the best FVD (198.99), PSNR (19.18), and temporal consistency (95.15 SC / 95.21 BC). The unmasked image branch improves FVD without improving Dice, illustrating a mismatch between global quality and control adherence. Masking improves both at the same three-step budget, supporting spatially selective teacher guidance.

\begin{table}[!htbp]
\centering\scriptsize
\setlength{\tabcolsep}{1.5pt}
\renewcommand{\arraystretch}{1.05}
\caption{\textbf{Stage-III distillation ablation.} All variants share 50 full-hand HandBench cases (hands in every frame) and three-step sampling. $\checkmark(\text{BG})$ excludes hand regions from image guidance. \textbf{Best}/\underline{second-best}.}
\label{tab:stage3_ablation}
\begin{adjustbox}{max width=\linewidth}
\begin{tabular}{cccrrrrr@{}}
\toprule
\multicolumn{3}{c}{Distillation Config.} & \multicolumn{5}{c}{Metrics} \\
\cmidrule(lr){1-3}\cmidrule(lr){4-8}
\shortstack{Video\\Score} & \shortstack{Image\\Score} & \shortstack{Video\\Shift} & FVD $\downarrow$ & \shortstack{MANO\\Dice $\uparrow$} & PSNR $\uparrow$ & SC $\uparrow$ & BC $\uparrow$ \\
\midrule
$\times$ & $\times$ & -- & 404.68 & 89.60 & 18.32 & 91.06 & 93.53 \\
$\checkmark$ & $\times$ & 1 & 303.61 & 90.53 & 17.20 & 93.62 & 94.54 \\
$\checkmark$ & $\times$ & 12 & 289.05 & \underline{92.03} & 18.02 & 94.03 & 94.46 \\
$\checkmark$ & $\checkmark$ & 12 & \underline{208.75} & 92.00 & \underline{18.89} & \underline{94.79} & \underline{94.96} \\
$\checkmark$ & $\checkmark(\text{BG})$ & 12 & \textbf{198.99} & \textbf{92.50} & \textbf{19.18} & \textbf{95.15} & \textbf{95.21} \\
\bottomrule
\end{tabular}
\end{adjustbox}
\end{table}

\section{Conclusion}
\label{sec:conclusion}

We presented LynnReal-Hand, an end-to-end framework for low-latency, hand-controlled video world modeling. Our 100K interaction corpus and HandBench provide aligned supervision and evaluation across diverse scenes. Interaction-Preserving Training combines local hand supervision with region-aware causal distillation, while the cache-free Ultra-Fast-VAE removes the remaining decoding bottleneck. Together, they preserve fine manipulation details at 18.03\,FPS on a single H200 GPU. We will release all data, benchmarks, weights, and code for interactive world modeling.

\textbf{Limitations and future work.} Our local two-chunk causal window ($\mathrm{W2}$) prioritizes streaming latency but lacks explicit long-term 3D memory. Persistent spatial memory could improve object permanence when objects leave and re-enter the view. Coupling kinematic controls with physical contact priors for closed-loop dexterous robot learning is an important direction for future work.

\clearpage
{\small
\bibliographystyle{unsrtnat}
\bibliography{main}
}

\clearpage
\appendix
\raggedbottom
\setlength{\parskip}{0pt}
\linespread{1.06}\selectfont
\titlespacing*{\section}{0pt}{12pt plus 2pt minus 2pt}{6pt plus 1pt minus 1pt}
\titlespacing*{\subsection}{0pt}{8pt plus 2pt minus 2pt}{4pt plus 1pt minus 1pt}
\phantomsection
\label{sec:supp_contents}
\pdfbookmark[0]{Appendix}{report.appendix}
\begin{center}
{\Large\bfseries Appendix}\par\vspace{5pt}
{\large LynnReal-Hand}
\end{center}

\setlength{\textfloatsep}{10pt plus 2pt minus 2pt}
\setlength{\floatsep}{8pt plus 2pt minus 2pt}
\setlength{\dbltextfloatsep}{10pt plus 2pt minus 2pt}
\setlength{\dblfloatsep}{8pt plus 2pt minus 2pt}
\renewcommand{\topfraction}{.92}
\renewcommand{\dbltopfraction}{.92}
\renewcommand{\textfraction}{.06}
\renewcommand{\floatpagefraction}{.88}
\renewcommand{\dblfloatpagefraction}{.88}
\captionsetup{font=small,skip=4pt}

\section{Evaluation Metrics}
\label{sec:supp_metrics}

\subsection{Video Generation Metrics}
\label{sec:report_generation_metrics}

\textbf{Alignment and aggregation.}
All paired metrics are evaluated on common ground-truth frame indices without temporal interpolation. Scalar metrics are first averaged across sequences within each data source and then macro-averaged with equal weights across sources, whereas FVD is computed over pooled video features.

\textbf{Hand control accuracy.}
To measure kinematic adherence, a frozen WiLoR detector~\cite{wilor} reconstructs 3D hand meshes independently from generated frames. Both reference and predicted MANO~\cite{mano} meshes are projected onto the shared $1344\times768$ evaluation canvas. Let $M_t^R$ and $M_t^G$ denote the union binary silhouettes of left and right hands at frame $t$. We compute the projected mesh-silhouette Dice score as
\begin{equation}
 D_{\mathrm{MANO}} = \frac{100}{|\mathcal{T}_v|} \sum_{t \in \mathcal{T}_v}
 \frac{2|M_t^R \cap M_t^G|}{|M_t^R| + |M_t^G|},
\end{equation}
where $\mathcal{T}_v = \{t : |M_t^R| > 0\}$ denotes the set of frames with valid reference hands.
Evaluation is conducted at 4\,Hz on aligned source timestamps without mask dilation or post-hoc pose alignment. Frames where the generator fails to produce a detectable hand receive a score of zero, while frames lacking ground-truth hands are excluded from $\mathcal{T}_v$.

\textbf{Visual quality and temporal consistency.}
PSNR and grayscale SSIM~\cite{ssim} are computed on an aspect-preserving $1344\times768$ canvas (including letterbox padding); PSNR pools mean squared error over each video prior to taking the logarithm, and SSIM uses an $11\times11$ Gaussian kernel ($\sigma=1.5$) excluding the 5-pixel border. Perceptual similarity is measured via VGG-based LPIPS~\cite{vgg,lpips} at $448\times256$ resolution with inputs normalized to $[-1,1]$. FVD~\cite{fvd} compares the mean and covariance of 400-dimensional Kinetics-I3D~\cite{i3d} features extracted from 32 uniformly sampled frames per clip. For semantic and temporal coherence, let $u_t$, $v_t$, and $q$ denote the unit-normalized CLIP ViT-B/32~\cite{clip} embeddings of the generated frame, reference frame, and text prompt, respectively. Weighted CLIP similarity is defined as
\begin{equation}
 C_{\mathrm{CLIP}}=100\!\left(
 \frac{0.2}{T}\sum_{t=1}^{T}u_t^\top q
 +\frac{0.2}{T-1}\sum_{t=1}^{T-1}u_t^\top u_{t-1}
 +\frac{0.6}{T}\sum_{t=1}^{T}u_t^\top v_t
 \right).
 \label{eq:weighted_clip}
\end{equation}
evaluated on $640\times360$ frames via the standard $224\times224$ center-crop processor. Following VBench~\cite{vbench}, Subject Consistency (SC) and Background Consistency (BC) average non-negative cosine similarities to the first and immediately preceding frames using DINO ViT-B/16~\cite{dino} and CLIP ViT-B/32 features, respectively, scaled to $[0,100]$.

\textbf{Camera adherence (Cam-BG).}
Because monocular SLAM can fail on short or low-texture clips, we evaluate camera adherence through hand-masked background motion agreement at 4\,Hz and $640\times360$ resolution. Specifically, hand silhouettes from adjacent frames are merged via convex hull and dilated with a $31\times31$ kernel, excluding a 12-pixel image border and letterbox padding. Shi--Tomasi corners tracked via Lucas--Kanade optical flow (with forward--backward error $\leq 1$\,px) are used to fit RANSAC homographies, requiring at least 40 valid tracks, 24 inliers ($\geq 60\%$), and spatial coverage across at least three image quadrants. For reference and generated displacement vectors $a$ and $b$ on a uniform grid, let $c(a,b)$ denote their non-negative cosine similarity and $r(a,b) = \min(\|a\|,\|b\|)/\max(\|a\|,\|b\|,10^{-8})$ their magnitude ratio. With threshold $\tau=0.5$\,px, point-wise agreement is given by
\begin{equation}
 A(a,b)=
 \begin{cases}
 c(a,b)r(a,b), & \|a\|\geq\tau,\\
 \exp\!\left(-(\|b\|/\tau)^2\right), & \|a\|<\tau.
 \end{cases}
\end{equation}
Clip scores average $100A(a,b)$ over valid grid points and eligible frame pairs (requiring $\geq 25\%$ observable reference pairs per video), assigning zero to frame pairs where homography estimation fails on the generated output.

\subsection{Decoder Reconstruction Metrics and Timing}
\label{sec:supp_vae_evaluation}
\textbf{Benchmark protocol.}
All VAE baselines in Table~\ref{tab:decoder} are evaluated on an identical held-out set of 200 videos (29,818 frames, disjoint from HandBench) encoded with the native H3 encoder at $1344\times768$. Global PSNR pools squared pixel errors across all frames prior to taking the logarithm. Temporal error ($\text{deltaMAE}$) averages $|e_t - e_{t-1}|$ for normalized frame error $e_t = (\widehat{I}_t - I_t)/255$, and LPIPS is evaluated at frames $\{0, 8, 16\}$ per 17-frame chunk at $448\times256$.

\textbf{Latency measurement and precision.} Latency is measured on a single H200 GPU over 501 chunks after warmup (including tile stitching and host synchronization). Comparing numerical precision on the same 200-video cohort, FP16 and W8A8 Ultra-Fast-VAE achieve 34.45\,dB and 34.43\,dB PSNR, respectively, confirming negligible quantization degradation ($0.02$\,dB).

\section{Data Engine Construction}
\label{sec:supp_data_engine}

To train a generalizable hand-controlled world model, we construct a unified curation and annotation pipeline that converts heterogeneous first-person videos into 100K synchronized multimodal interaction clips.

\textbf{Video collection and quality filtering.}
We aggregate egocentric recordings from public datasets and web sources alongside synthetic first-person interactions rendered by MiniMax-H3~\cite{minimaxh3}. Candidate videos must have a resolution of at least 720p, a frame rate $\geq 24$\,fps, and $>200$ frames (retaining only egocentric views from multi-view datasets). Shot boundaries are detected with OmniShotCut~\cite{omnishotcut} to discard transition artifacts, and continuous shots are sliced into 125- or 240-frame clips at 24\,fps. To ensure a crisp conditioning frame, we select the starting frame with the highest grayscale Laplacian variance within the first ten frames of each candidate cut. All clips are standardized to H.264 with a 768-pixel short side while preserving aspect ratio, and a low-resolution subset is enhanced via SparkVSR~\cite{yu2026sparkvsr}. We filter out poorly exposed clips via luminance histograms, retain the top 80\% ranked by the COVER perceptual quality score~\cite{cover}, and remove persistent lower-third subtitles detected by PaddleOCR (discarding subtitled landscape clips and cropping portrait clips only when subtitle removal trims $\leq 25\%$ of image height).

\textbf{3D hand annotation and trajectory completion.}
Per-frame hand bounding boxes are detected via WiLoR~\cite{wilor}; duplicate or temporally inconsistent detections are pruned, and clips with hand presence below 50\% are discarded. Short detection gaps bounded by same-side tracks are linearly interpolated when the hand lies away from frame boundaries, while longer occlusions are recovered using SAM~3~\cite{sam3} prompted within an expanded search region around the preceding detection. After temporal smoothing, clips with $<90\%$ hand coverage are rejected. WiLoR then regresses MANO~\cite{mano} pose, shape, global orientation, and camera-space translation from the completed tracks, which are temporally smoothed and rasterized onto a black background at native resolution and frame rate.

\textbf{Camera pose estimation.}
Camera intrinsics and per-frame extrinsics are estimated using ViPE~\cite{vipe}, initialized with GeoCalib intrinsics and constrained by MoGe-2 metric depth priors. To prevent dynamic hands from corrupting static scene bundle adjustment, we dilate the rasterized hand silhouettes via morphological closing and mask them out from geometric residuals. Trajectories exhibiting missing poses, frame-index mismatches, translation spikes ($>5\times$ local mean), abrupt inter-frame rotations ($>60^\circ$), or cumulative rotations exceeding $100^\circ$ within 1.2\,s are automatically rejected.

\textbf{Semantic captioning and interaction verification.}
We employ Qwen3.8-Flash-Next~\cite{qwen38} (served via vLLM) on 1\,Hz keyframes to generate scene descriptions summarizing initial object states and visible hand--object interactions, explicitly excluding camera motion descriptions so viewpoint control remains disentangled in the camera condition. In a secondary verification pass sampled every 25 frames, the VLM filters out clips that violate any of three criteria: (i) strictly egocentric viewpoint, (ii) presence of only the camera wearer's hands without third-party actors, and (iii) purposeful hand--object interaction (e.g., grasping, placing, assembly, or tool manipulation, excluding idle hovering or empty gestures).

\textbf{Diversity-balanced sampling.}
Finally, we sample 100K clips without replacement using weights inversely proportional to the joint frequency of VLM-assigned background and activity categories, upweighting rare environments and manipulation skills.

\section{Training and Implementation Details}
\label{sec:supp_method}

\subsection{Stage-I Hand-Aware Pretraining}
\label{sec:supp_stage1_training}
The 33B bidirectional DiT backbone is trained in BF16 with FlashAttention-3~\cite{flashattn3} while keeping all conditioning encoders frozen. Qwen3-VL~\cite{qwen3vl} encodes the text prompt and initial frame $I_0$, while the pretrained H3 VAE~\cite{minimaxh3} encodes target and hand-conditioning videos into $24\times42$ spatial latent grids using $1\times2\times2$ patchification. Camera poses use fixed intrinsics and first-frame-relative extrinsics injected via PRoPE~\cite{prope}. Clean anchor and hand conditioning latents are perturbed with minimal noise ($0.999X + 0.001\epsilon$ at $\sigma=0.001$) for numerical stability. For pixel-level hand supervision, we sample a $16\times16$ latent crop centered on the hand mask (with a 0.2 spatial margin), decode 5 target latents alongside 2 detached context latents, and compute LPIPS~\cite{lpips} and DINOv2~\cite{dinov2} losses on the corresponding $256\times256$ hand region. Loss weights are set to $\lambda_R=5$, $\lambda_P=0.1$, and $\lambda_D=0.05$.

\textbf{Stage-I ablation evaluation.}
Table~\ref{tab:stage1_ablation} uses a fixed random sample of 100 cases from the complete 200-case HandBench, shared by all four loss configurations to reduce the cost of full-sequence inference. Its subset-level metrics are reported separately from the main and OOD benchmark cohorts in Table~\ref{tab:HandBench}.

\subsection{Stage-II Causal Flow-Map Adaptation}
\label{sec:supp_stage2_training}
Stage~II adapts the Stage-I model using rank-384 LoRA~\cite{lora} adapters and interval-conditioning embeddings (2.08B trainable parameters total). Optimization uses FP32 AdamW with $(\beta_1,\beta_2)=(0.9,0.95)$, cosine learning-rate decay, EMA decay $0.999$, and classifier-free guidance disabled.

To approximate the trajectory total derivative in Eq.~\eqref{eq:anyflow_loss}, let $d=\operatorname{sg}(v_\theta)$, $\sigma_-=\max(r,\sigma-h)$, and $\sigma_+=\min(1,\sigma+h)$. We compute the central finite difference
\begin{equation}
 \widehat D_\sigma U=\operatorname{sg}\!\left[\frac{U_+-U_-}{\sigma_+-\sigma_-}\right],
 \qquad U_\pm=U_\theta\!\left(z_{k,\sigma}+(\sigma_\pm-\sigma)d,\,\sigma_\pm,\,r\right).
 \label{eq:anyflow_deriv}
\end{equation}
where $U_\pm$ share identical conditioning inputs and gradients propagate solely through the interval prediction $U_\theta$. Interior timestep pairs $(r, \sigma)$ are sampled by applying identical monotone schedule shifts to $(vu, u)$ for independent uniform $u, v \sim \mathcal{U}(0,1)$, with the stop-gradient loss balancing defined below.

\textbf{Stop-gradient loss balancing.}
Let $\rho_i=U_\theta+(\sigma_i-r_i)\widehat D_{\sigma_i}U_\theta-v_i^\star$ and $\ell_i=\|\rho_i\|_M^2$ be the residual and masked mean-squared error for example $i$. Given the mean diagonal loss $\mu_{\mathrm{diag}}$ over examples with $r_i=\sigma_i$, non-diagonal examples are rescaled as
\begin{equation}
 \ell_i^{\mathrm{bal}}=
 \begin{cases}
  \ell_i, & r_i=\sigma_i,\\
  \operatorname{sg}\!\left[
   \dfrac{\mu_{\mathrm{diag}}}{\ell_i+\varepsilon_{\mathrm{bal}}}
  \right]\ell_i, & r_i<\sigma_i,
 \end{cases}
 \label{eq:anyflow_balance}
\end{equation}
where $\varepsilon_{\mathrm{bal}}>0$ prevents division by zero. The scale factor is detached, so gradients pass through the trainable interval prediction rather than through the balancing ratio. If no diagonal reference is available, the unscaled loss is used.

\subsection{Stage-III Region-Aware Score Distillation}
\label{sec:supp_stage3_training}
The differentiable latent-to-image bridge $B$ maps 5 native 24-channel H3 latent planes through a width-256 projection, eight spatial residual blocks, and $2\times$ pixel-shuffle upsampling, conditioned on physical frame-offset embeddings to extract individual frame latents at 4 temporal positions per rollout. After pre-fitting $B$ on paired video--image latents, its weights are frozen while gradients flow through $J_B^\top$ into the video rollout. For the image branch, a frozen Qwen-Image~\cite{qwenimage} teacher ($\text{CFG}=4.0$ with per-image L2 norm rescaling, unconditioned on hands or camera) and a trainable Wan2.1-T2V-1.3B~\cite{wan} single-frame critic evaluate corrupted bridge outputs; following VJP backpropagation through $B$, image gradients inside the dilated hand mask $M_H$ are zeroed out.

\textbf{Per-image L2 norm rescaling.}
Let $f^c,f^u\in\mathbb R^{B\times C\times H\times W}$ be the image teacher's conditional and unconditional flow predictions. For each batch item $b$, the wrapper applies CFG with scale $4.0$ and rescales the resulting flow:
\begin{equation}
 f_b^{\mathrm{cfg}}=f_b^u+4(f_b^c-f_b^u),\quad
 r_b=\frac{\max(\|f_b^c\|_{C,H,W},10^{-6})}
              {\max(\|f_b^{\mathrm{cfg}}\|_{C,H,W},10^{-6})},\quad
 f_b^{\mathrm{guided}}=r_bf_b^{\mathrm{cfg}}.
 \label{eq:image_norm_rescaling}
\end{equation}
Here $\|\cdot\|_{C,H,W}$ is the joint L2 norm over all channels and spatial positions of one image. Each image receives one scale factor; batch items are reduced independently. These are the reduction dimensions of our teacher wrapper.

For the video branch, both the frozen Stage-I teacher and trainable video critic operate with $\text{CFG}=3.0$ (where the unconditional branch retains initial-image and camera conditions while dropping hand and text prompts). Student rollouts use a shift-3 schedule, whereas video critic/teacher evaluations use a shift-12 noise warp ($\gamma > 1$) to emphasize hand kinematics. Both critics receive 5 optimizer steps per generator update after a 10-step warmup, during which the image guidance weight $\lambda_i$ ramps linearly from $0.001$ to $0.1$, with student EMA decay $0.999$ and detached rollout histories.

\textbf{Ablation evaluation.}
The Stage-III configurations in Table~\ref{tab:stage3_ablation} are evaluated on the same 50-case \emph{full-hand} subset selected from HandBench. Each selected sequence has hands visible in every frame, providing continuous hand-control targets for a fair comparison. All configurations use the same three-step causal sampling budget and the metrics defined in Appendix~\ref{sec:supp_metrics}.

\subsection{Causal Attention Masks and Open-Ended Autoregression}
\label{sec:supp_attention_rope}
Across the three training stages, attention visibility transitions from full-sequence bidirectional attention (Stage~I) to two-chunk causal teacher forcing (Stage~II) and self-rollout autoregression (Stage~III).

\textbf{Two-chunk causal teacher forcing.}
As shown in Fig.~\ref{fig:pipeline}(b), Stage~II processes paired streams of clean ground-truth chunks $\mathcal{Y}_k$ and noisy target chunks $\mathcal{X}_k$. Let $\mathcal{P}$ denote the shared text/initial-image prefix and independent image anchor, and let $\mathcal{H}_k$ denote the aligned hand-condition tokens for chunk $k$. Defining $\mathcal{A}(Q)$ as the set of key groups visible to query group $Q$, we enforce
\begin{equation}
\begin{aligned}
 \mathcal{A}(\mathcal{Y}_k) &= \mathcal{P} \cup \mathcal{H}_k \cup \mathcal{Y}_{k-1} \cup \mathcal{Y}_k, \\
 \mathcal{A}(\mathcal{X}_k) &= \mathcal{P} \cup \mathcal{H}_k \cup \mathcal{Y}_{k-1} \cup \mathcal{X}_k,
\end{aligned}
\label{eq:teacher_forcing_attention}
\end{equation}
with $\mathcal{Y}_{-1}=\emptyset$. Noisy queries are strictly masked from their paired clean targets $\mathcal{Y}_k$, other noisy chunks, and future chunks. To prevent indirect target leakage through conditioning tokens, prefix queries in $\mathcal{P}$ attend only within $\mathcal{P}$ and each hand group $\mathcal{H}_k$ attends only to itself. Within each visible block, full spatial-temporal attention is preserved; causality operates at the chunk level under a two-chunk window ($\mathrm{W2}$, comprising the current chunk and one full-resolution preceding chunk).

\textbf{Causal self-rollout.}
In Stage~III (Fig.~\ref{fig:pipeline}(c)), we eliminate exposure bias by replacing clean history $\mathcal{Y}_{k-1}$ with the student's completed, gradient-detached prediction $\widehat{\mathcal{Y}}_{k-1}$:
\begin{equation}
 \mathcal{A}(\mathcal{X}_k) = \mathcal{P} \cup \mathcal{H}_k \cup \widehat{\mathcal{Y}}_{k-1} \cup \mathcal{X}_k.
 \label{eq:self_rollout_attention}
\end{equation}
During the three denoising steps of chunk $k$, keys and values for $\mathcal{P}$, $\mathcal{H}_k$, and $\widehat{\mathcal{Y}}_{k-1}$ are cached. Once chunk $k$ is denoised, a single clean forward pass commits its KV states to serve as history for chunk $k+1$, while older chunks ($<k$) are evicted. Meanwhile, the bidirectional video teacher and critic evaluate corrupted rollouts $Z_\theta$ across the full sequence span, and the image branch evaluates individual decoded frames.

\textbf{Local RoPE reset for open-ended inference.}
\label{sec:streaming_rope}
While a $\mathrm{W2}$ sliding window bounds KV cache memory, unbounded physical frame indices would extrapolate rotary positional embeddings (RoPE) beyond the training horizon during long rollouts. We therefore maintain a bounded local temporal clock in the Stage-III student. For physical chunk index $k$, window size $W$, and oldest retained chunk $a_k = \max(0, k - W + 1)$, let $h_u$ denote the base temporal coordinate of token $u$ within a single chunk template and $\Delta = 17 \times 5/3$ denote the native chunk span (5 latent planes corresponding to 17 video frames). Hand tokens $\mathcal{H}_k$ always reuse $h_u$, whereas each retained video chunk $j \in [a_k, k]$ is assigned local temporal coordinates:
\begin{equation}
 p_{j,u}^{(k)} = h_u + (j - a_k + 1)\Delta, \qquad a_k \leq j \leq k.
 \label{eq:local_rope_positions}
\end{equation}
Prefix tokens $\mathcal{P}$ retain fixed coordinates. When the window advances from $[k-1, k]$ to $[k, k+1]$, chunk $k$ shifts from the second local video slot to the first. To keep relative coordinates consistent without compounding phase rotations, we cache unrotated keys $K_j^{\mathrm{raw}}$ at every DiT layer and re-apply RoPE using the updated local positions:
\begin{equation}
 \widetilde{K}_j^{(k)} = R\!\left(p_j^{(k)}\right) K_j^{\mathrm{raw}},
 \label{eq:local_rope_keys}
\end{equation}
where $R(\cdot)$ incorporates both updated temporal positions and invariant spatial coordinates. Camera PRoPE continues to use physical frame indices and poses independently of this local clock reset, enabling open-ended autoregressive generation with bounded memory and constant per-chunk latency.

\subsection{Ultra-Fast-VAE Training Details}
\label{sec:vae_training_details}
Table~\ref{tab:vae_losses} gives the weights for Eq.~\eqref{eq:vae_objective}. All components use mean reduction; a combined entry has one outer weight.

\begin{table}[t]
\centering\scriptsize
\caption{\textbf{Training loss components and weights for Ultra-Fast-VAE.} All weights scale mean-reduced loss terms combining pixel-level fidelity, hand-localized detail, frozen-teacher operator/feature distillation, and intra-/inter-chunk temporal smoothness.}
\label{tab:vae_losses}
\begin{adjustbox}{max width=\linewidth}
\begin{tabular}{lrlr}
\toprule
Loss term & Weight & Loss term & Weight \\
\midrule
Dense image MSE & 50 & Sampled teacher MSE & 10 \\
Hand-weighted L1 +5 MSE & 0.05 & Image Charbonnier +5 dense MSE & 0.25 \\
Average-pool pyramid (2,4,8) & 0.05 & Spatial error gradients & 0.02 \\
High-pass detail & 0.03 & Haar coefficients & 0.02 \\
Haar high-frequency energy & 0.005 & 16-pixel patch-boundary error & 0.02 \\
Spatial patch-phase bias & 0.01 & SSIM loss & 0.02 \\
Hand-weighted temporal delta & 0.10 & Temporal acceleration & 0.02 \\
Four-phase temporal bias & 0.02 & 17-frame chunk-boundary delta & 0.05 \\
Local branch + normalized last-three features & 0.005 & DINOv2 feature L1 & 0.001 \\
VGG LPIPS & 0.005 & Native blended tile-seam error & 0.005 \\
\bottomrule
\end{tabular}
\end{adjustbox}
\end{table}

\textbf{Pixel reconstruction and teacher guidance.}
Dense RGB MSE and sampled full-width teacher MSE have weights 50 and 10. Teacher and student use matched sampled spatial views. The remaining terms regularize spatial detail, temporal continuity, and hidden representations.

\textbf{Spatial reconstruction terms.}
Hand-weighted L1 plus five times MSE combines absolute and squared pixel errors, with greater weight on hand regions. The Charbonnier term combines a smoothed absolute error with five times dense MSE. 
Average-pool pyramid supervision compares images after pooling by factors of two, four, and eight. Spatial-gradient supervision matches changes between neighboring pixels in both image directions. High-pass supervision compares fine image detail, while the Haar terms match wavelet coefficients and high-frequency energy. The 16-pixel patch-boundary term penalizes reconstruction errors along the readout patch grid. SSIM supervision encourages structural similarity to the reference image.

\textbf{Temporal reconstruction terms.}
Temporal-delta supervision matches the change between adjacent reconstructed frames to the corresponding ground-truth change, with additional weight on regions occupied by hands in either frame. Temporal acceleration compares how these changes evolve over three consecutive frames. The chunk-boundary term applies temporal-delta supervision specifically to adjacent frames crossing a 17-frame chunk boundary. Ground-truth motion supplies the target, preserving rapid hand movement while reducing reconstruction flicker.

\textbf{Phase and tile-seam supervision.}
For the following three terms, RGB values are normalized to the range from zero to one, and prediction errors are computed in FP32. Spatial and temporal phase losses use 256-by-256 crops aligned to the 16-pixel grid. Tile-seam supervision uses the full native canvas after blending. These three terms use uniform averaging without additional hand-mask weighting.

\textbf{Spatial patch-phase bias (weight 0.01).}
Pixels are grouped by their horizontal and vertical offsets within each 16-by-16 patch. For each RGB channel, we average the reconstruction error for every offset across samples, frames, and patch locations. We subtract that channel's overall mean error, then average the absolute deviations across all offsets and channels. Subtracting the channel mean isolates periodic patch bias.

\textbf{Four-phase temporal bias (weight 0.02).}
Each 17-frame chunk is handled separately. We skip frame zero and form four groups: frames 1, 5, 9, and 13; frames 2, 6, 10, and 14; frames 3, 7, 11, and 15; and frames 4, 8, 12, and 16. For each sample, chunk, and RGB channel, we average the reconstruction error over the frames and spatial positions in each group. The loss measures the mean absolute deviation of these four group averages from their shared mean, discouraging phase-dependent temporal bias.

\textbf{Native blended tile-seam error (weight 0.005).}
After blending the decoded tiles, we inspect every non-initial tile's starting boundary and the end of its overlap region, counting each location once. At each boundary, we compare the predicted change between neighboring pixels with the ground-truth change. Horizontal and vertical errors are averaged separately and receive equal weight. This encourages seam transitions to match the reference image on the final blended canvas.

\textbf{Operator and perceptual distillation.}
Frozen full-width attention and feed-forward operators are evaluated on the compact decoder's hidden states, providing local targets for its compressed branches. The same table entry also aligns normalized features from the final three Transformer blocks. DINOv2~\cite{dinov2} compares corresponding image-patch features using absolute differences, while VGG LPIPS~\cite{lpips} provides learned perceptual supervision on matched views.

\textbf{Optimization and W8A8 export.}
We use FP32 AdamW with momentum coefficients 0.9 and 0.99, zero weight decay, and gradient clipping at 1.0. The learning rate is 0.00001 for attention and feed-forward branches and 0.000003 for the remaining parameters. Six late-stage checkpoints are averaged for export.
Deployment uses SmoothQuant~\cite{smoothquant} activation smoothing with strength 0.75, calibrated only on training clips. Linear projections use W8A8 quantization with per-token activation scaling, per-channel weight scaling, INT32 accumulation, and FP16 output. FlashAttention-3, fused kernels, and CUDA graphs provide further runtime acceleration.

\section{Inference and Runtime Protocols}
\label{sec:supp_timing}

\subsection{Evaluated Model Configurations}
\label{sec:supp_inference_variants}
\texttt{LynnReal-Hand-Base} generates at $1344\times768$ using joint $\text{CFG}=3.0$, a shift-3 Euler schedule with 30 noise levels and 29 updates, and the native H3 VAE. \texttt{LynnReal-Hand-Fast} runs at 540p using three shift-3 flow-map updates between four noise levels, $\mathrm{W2}$ history, Tiny-H3~\cite{taehv} hand encoding, and FP16 Ultra-Fast-VAE for quality benchmarks (and W8A8 Ultra-Fast-VAE for streaming deployment).
All training stages use the native H3 hand encoder. Fast deployment replaces it with Tiny-H3 through a compatible H3 latent interface to reduce online condition-encoding latency.

For baseline comparisons, JointControlVideo~\cite{jointcontrolvideo} and streaming Hand2World~\cite{hand2world} use their official released checkpoints. To ensure fair domain alignment, both bidirectional Hand2World and Generated Reality~\cite{generatedreality} are fine-tuned on our full 100K training corpus: bidirectional Hand2World adapts rank-256 attention LoRA and camera adapters while retaining its native 50-step sampler ($\text{CFG}=1.0$) and Wan VAE without TeaCache~\cite{teacache}, whereas Generated Reality is fine-tuned via rank-32 LoRA and motion encoder adaptation. All methods are evaluated at their native frame rates without temporal interpolation.

\textbf{Schedule-specific AdaLN caching.}
\label{sec:supp_adaln_memory}
Because per-block Adaptive LayerNorm (AdaLN) modulation vectors depend solely on the deterministic timestep schedule, we precompute and store only the modulation tensors required at the active sampling transitions and clean-conditioning boundaries, discarding the learned AdaLN linear projection weights at inference. This reduces parameter storage overhead from 535.3\,MiB (Base) to 55.4\,MiB (three-step Fast) without altering numerical outputs.

\subsection{Hardware and Runtime Measurement}
\label{sec:supp_timing_protocol}
All runtime measurements in Table~\ref{tab:generation_efficiency} use one NVIDIA H200 GPU. Offline latency includes condition encoding, denoising, and VAE decoding at 768p; the 33B Base model uses sequential device offloading. Hand2World (streaming) generates four-frame chunks at 768p. LynnReal-Hand-Fast runs at 540p with 0.741\,s latency and 18.03\,FPS throughput.

\subsection{Optimized Streaming Deployment}
\label{sec:supp_fast_timing}
LynnReal-Hand-Fast uses W8A8 DiT projections, FlashAttention-3, fused RMSNorm/RoPE/SwiGLU kernels, CUDA graphs over all 50 DiT blocks, and W8A8 Ultra-Fast-VAE. The internal canvas is $960\times544$, cropped to $960\times540$. Measurements follow a full warmup rollout. Peak memory is recorded with PyTorch's max-allocated memory tracker during steady-state rollouts, with initial-image encoders offloaded.

\textbf{DiT precision.}
The 192 quantized projections comprise fused QKV, attention output, and FFN up/down in blocks 1--48. They use signed INT8 weights with per-output-channel scales and signed INT8 activations with dynamic per-token scales. Quantization is symmetric over $[-127,127]$ with nearest-even rounding; GEMM accumulates in INT32, followed by dequantization, bias addition, and BF16 output.

Blocks 0 and 49 remain BF16, and input/output projections remain FP32. The non-accelerated DiT uses BF16 blocks with the same FP32 input/output projections. Ultra-Fast-VAE uses a separate W8A8 path with FP16 output (Appendix~\ref{sec:vae_training_details}).

\section{Additional Generator Analyses}
\label{sec:supp_generator_analysis}

\subsection{Score-Direction Diagnostics}
\label{sec:supp_score_probe}
Figure~\ref{fig:score_responses} probes denoised-prediction differences $\Delta_b = D_b^f - D_b^r$ on the same student rollout chunk before DMD normalization and hand masking. At each fixed noise level, these are proportional to the fake-minus-real score difference (Eq.~\eqref{eq:distillation_directions}). Video probes use low ($\sigma_v=0.25$) and high ($\sigma_v=0.90$) noise levels with $\text{CFG}=3.0$ for the real model and $\text{CFG}=1.0$ for the fake model. The image probe uses bridge-projected latents $B(Z_\theta)$ at $\sigma_i=0.638$, with $\text{CFG}=4.0$ for the real model and $\text{CFG}=1.0$ for the fake model. Heatmaps display channel-RMS prediction differences scaled to each branch's 99th percentile.

\needspace{8\baselineskip}
\section{Contributors}
\label{sec:contributors}

\noindent\textbf{Project leads:}
Shaohao Rui$^{1,2,3}$, Xiaofeng Mao$^{1,2,5}$.\par\smallskip
\noindent\textbf{Core contributors (equal contribution):}
Shaohao Rui$^{1,2,3}$, Chaoyang Zhang$^{1,3,4}$, Xiaofeng Mao$^{1,2,5}$.\par\smallskip
\noindent\textbf{Corresponding author:}
Weijie Ma$^{1,2}$.\par\medskip
\noindent\textbf{Affiliations:}
$^1$LynnReal AI;
$^2$Shanghai Innovation Institute;
$^3$Shanghai Jiao Tong University;
$^4$Xi'an Jiao Tong University;
$^5$Fudan University.

\end{document}